\def\year{2022}\relax
\documentclass[letterpaper]{article} 
\usepackage{aaai22}  
\usepackage{times}  
\usepackage{helvet}  
\usepackage{courier}  
\usepackage[hyphens]{url}  
\usepackage{graphicx} 
\usepackage{natbib}  
\usepackage{caption} 
\DeclareCaptionStyle{ruled}{labelfont=normalfont,labelsep=colon,strut=off} 
\usepackage{algorithm}
\usepackage{algorithmic}

\usepackage{newfloat}
\usepackage{listings}
\floatstyle{ruled}
\newfloat{listing}{tb}{lst}{}
\floatname{listing}{Listing}

\usepackage{color}
\usepackage{amsmath}
\usepackage{amssymb}
\usepackage{array}
\usepackage{hhline}
\usepackage{makecell}
\usepackage{tabularx}
\usepackage{multirow}
\usepackage{booktabs}
\usepackage[super]{nth}

\newcommand{\crc}[1]{{#1}}
\newcommand{\blue}[1]{{#1}}
 \newcommand{\tipadd}[1]{{#1}}
 \newcommand{\aaaiadd}[1]{{#1}}
\newcommand{\aaaidel}[1]{}

\DeclareFontFamily{OT1}{mathc}{}
\DeclareFontShape{OT1}{mathc}{m}{n}{ <-> mathc10 }{}

\newcolumntype{L}{>{\centering\arraybackslash}X}
\newcolumntype{R}{>{\raggedleft\arraybackslash}X}
\newcolumntype{C}{>{\arraybackslash}X}

\newcolumntype{Y}{>{\centering\arraybackslash}X}
\newcolumntype{Z}{>{\arraybackslash\hsize=1.3\hsize}X}
\newcolumntype{y}{>{\centering\arraybackslash\hsize=.8\hsize}X}
\newcolumntype{k}{>{\centering\arraybackslash\hsize=.3\hsize}X}
\newcolumntype{q}{>{\centering\arraybackslash\hsize=.5\hsize}X}
\newcolumntype{v}{>{\centering\arraybackslash\hsize=.4\hsize}X}
\newcolumntype{K}{>{\centering\arraybackslash\hsize=.5\hsize}X}
\newcolumntype{V}{>{\raggedleft\arraybackslash\hsize=.4\hsize}X}
\newcolumntype{s}{>{\hsize=.3\hsize}X}
\newcolumntype{G}{>{\hsize=0.7\hsize}X}
\newcolumntype{T}{>{\hsize=1.9\hsize}X}
\newcolumntype{t}{>{\hsize=1.9\hsize}X}
\newcolumntype{Q}{>{\centering\arraybackslash\hsize=0.53\hsize}X}
\newcolumntype{e}{>{\hsize=.9\hsize}X}

\newcolumntype{A}{>{\centering\arraybackslash\hsize=0.8\hsize}X}
\newcolumntype{S}{>{\centering\arraybackslash\hsize=0.9\hsize}X}
\newcolumntype{D}{>{\raggedright\arraybackslash\hsize=1.7\hsize}X}
\newcolumntype{F}{>{\raggedright\arraybackslash\hsize=2\hsize}X}

\newcommand{\tabincell}[2]{\begin{tabular}{@{}#1@{}}#2\end{tabular}}

\newcommand{\ie}{\emph{i.e}. }

\newcommand{\multiref}[2]{\ref{#1}--\ref{#2}} 

\usepackage{threeparttable}

\title{Active Boundary Loss for Semantic Segmentation}
\author {
    Chi Wang\textsuperscript{\rm 1,2},
    Yunke Zhang\textsuperscript{\rm 1,2},
    Miaomiao Cui\textsuperscript{\rm 2},
    Peiran Ren\textsuperscript{\rm 2} \\
    Yin Yang\textsuperscript{\rm 3},
    Xuansong Xie\textsuperscript{\rm 2},
    Xian-Sheng Hua\textsuperscript{\rm 2},
    Hujun Bao\textsuperscript{\rm 1},
    Weiwei Xu\textsuperscript{\rm 1}\thanks{Corresponding author.}
}
\affiliations {
    \textsuperscript{\rm 1} State~Key~Lab~of~CAD\&CG,~Zhejiang~University \quad     \textsuperscript{\rm 2} Alibaba~Inc \quad \textsuperscript{\rm 3} Clemson~University\\
    \{wangchi1995, yunkezhang\}@zju.edu.cn, \{miaomiao.cmm, peiran.rpr\}@alibaba-inc.com \\
    yin5@clemson.edu, xingtong.xxs@taobao.com, xiansheng.hxs@alibaba-inc.com, \{bao, xww\}@cad.zju.edu.cn
}

\usepackage{bibentry}

\begin{document}


\twocolumn[{%
\renewcommand\twocolumn[1][]{#1}%
\maketitle
\begin{figure}[H]
\hsize=\textwidth
    \centering
	\setlength{\tabcolsep}{0.5pt}
		\begin{tabular}{cccccc}
        \includegraphics[width=0.165\textwidth,keepaspectratio]{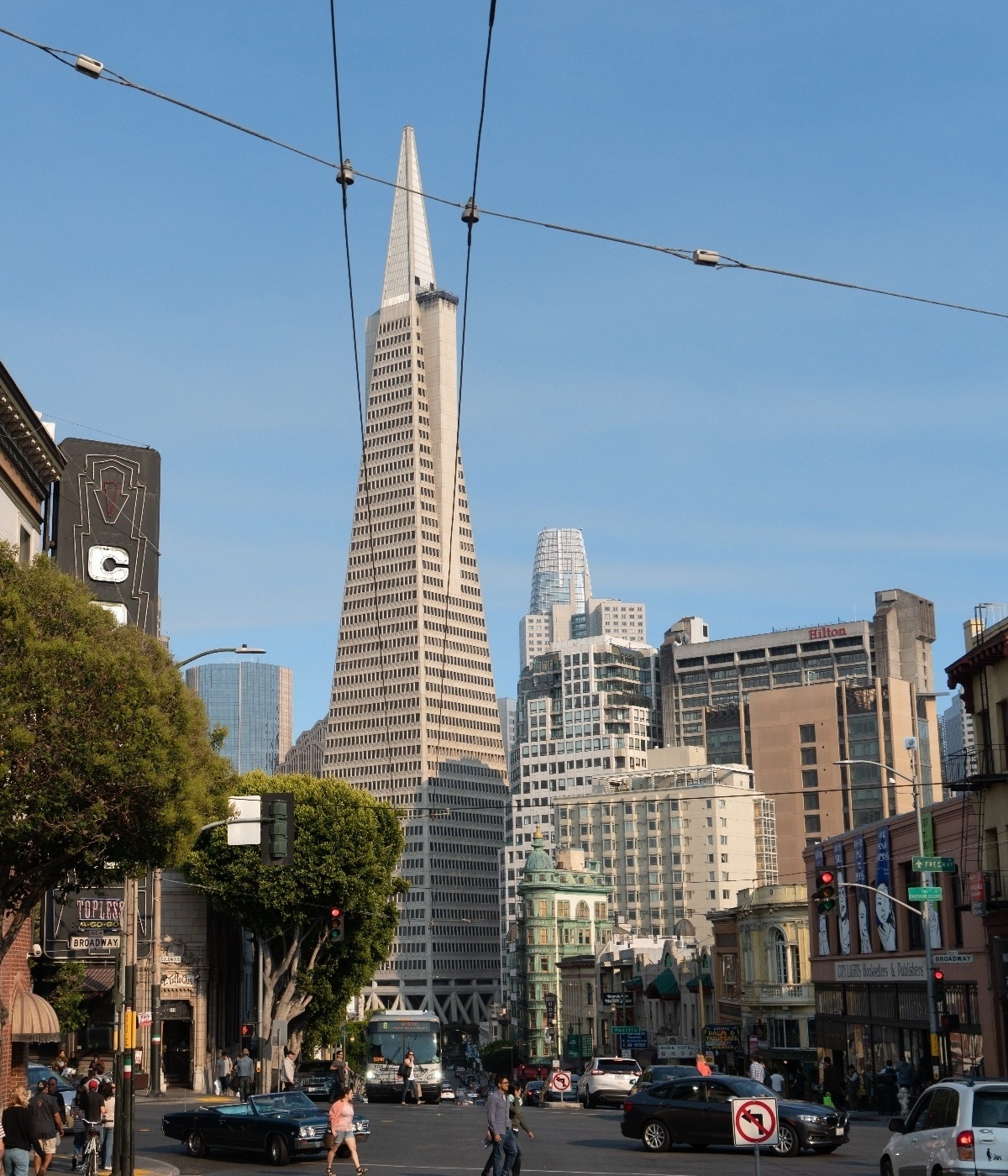} &
        \includegraphics[width=0.165\textwidth,keepaspectratio]{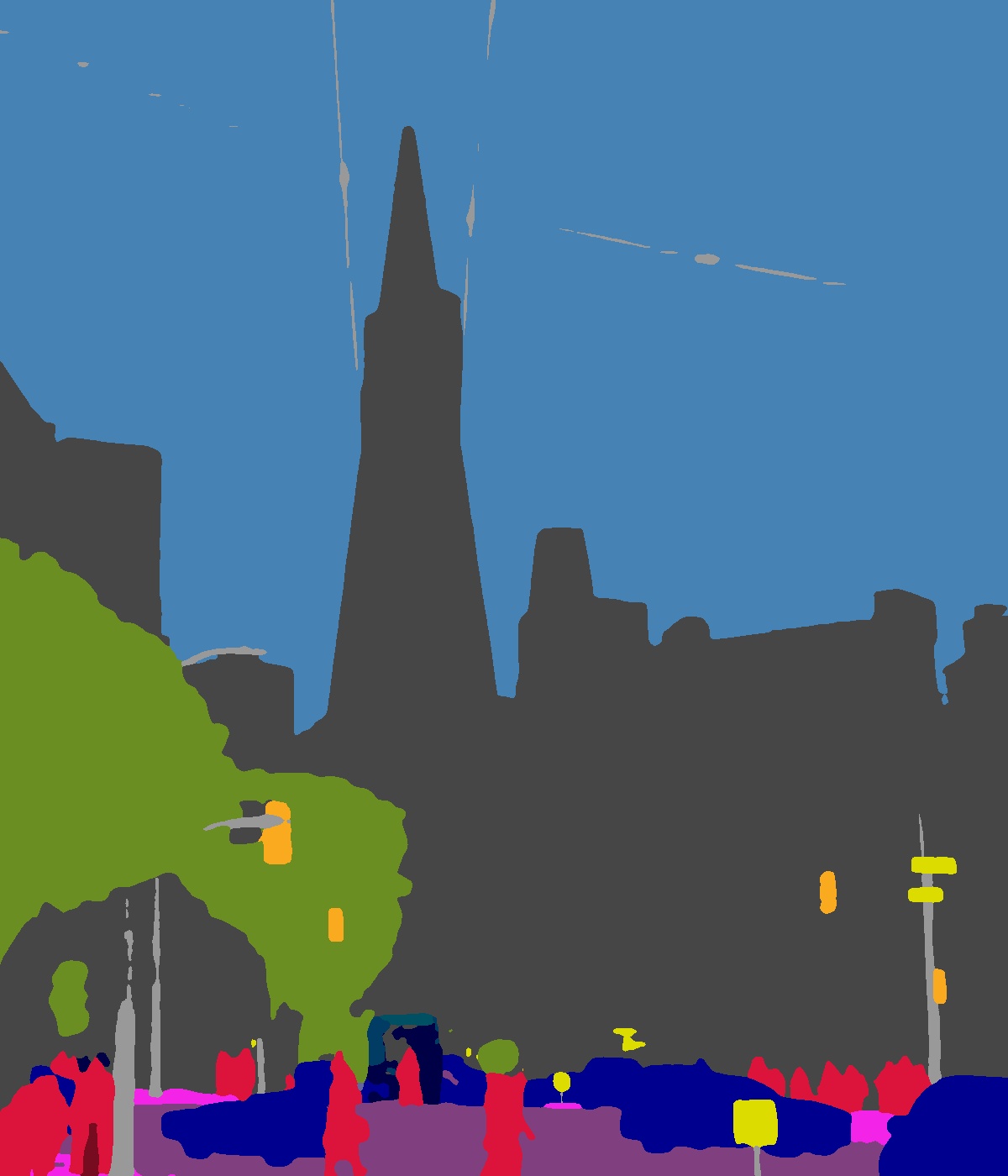}  &
        \includegraphics[width=0.165\textwidth,keepaspectratio]{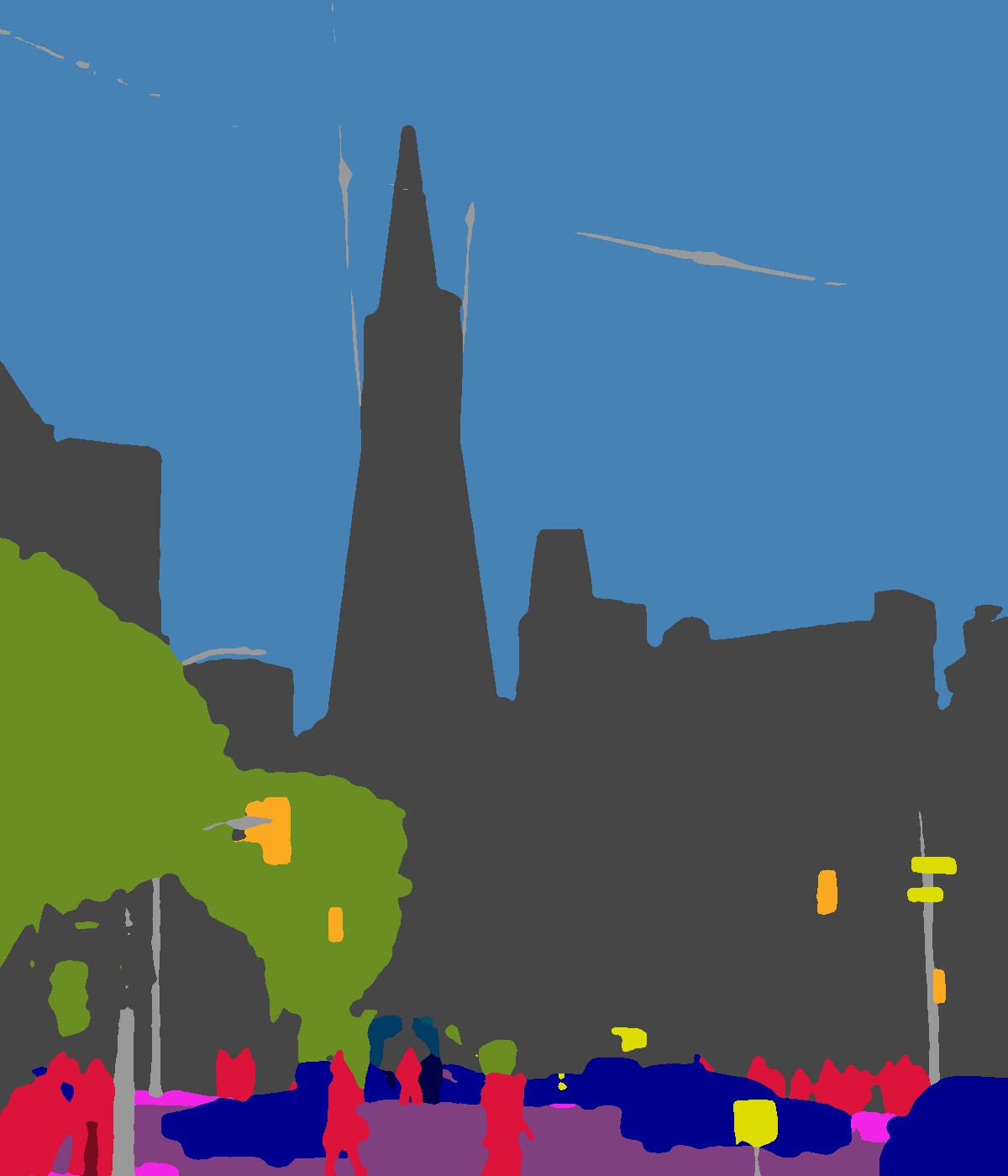} &
        \includegraphics[width=0.165\textwidth,keepaspectratio]{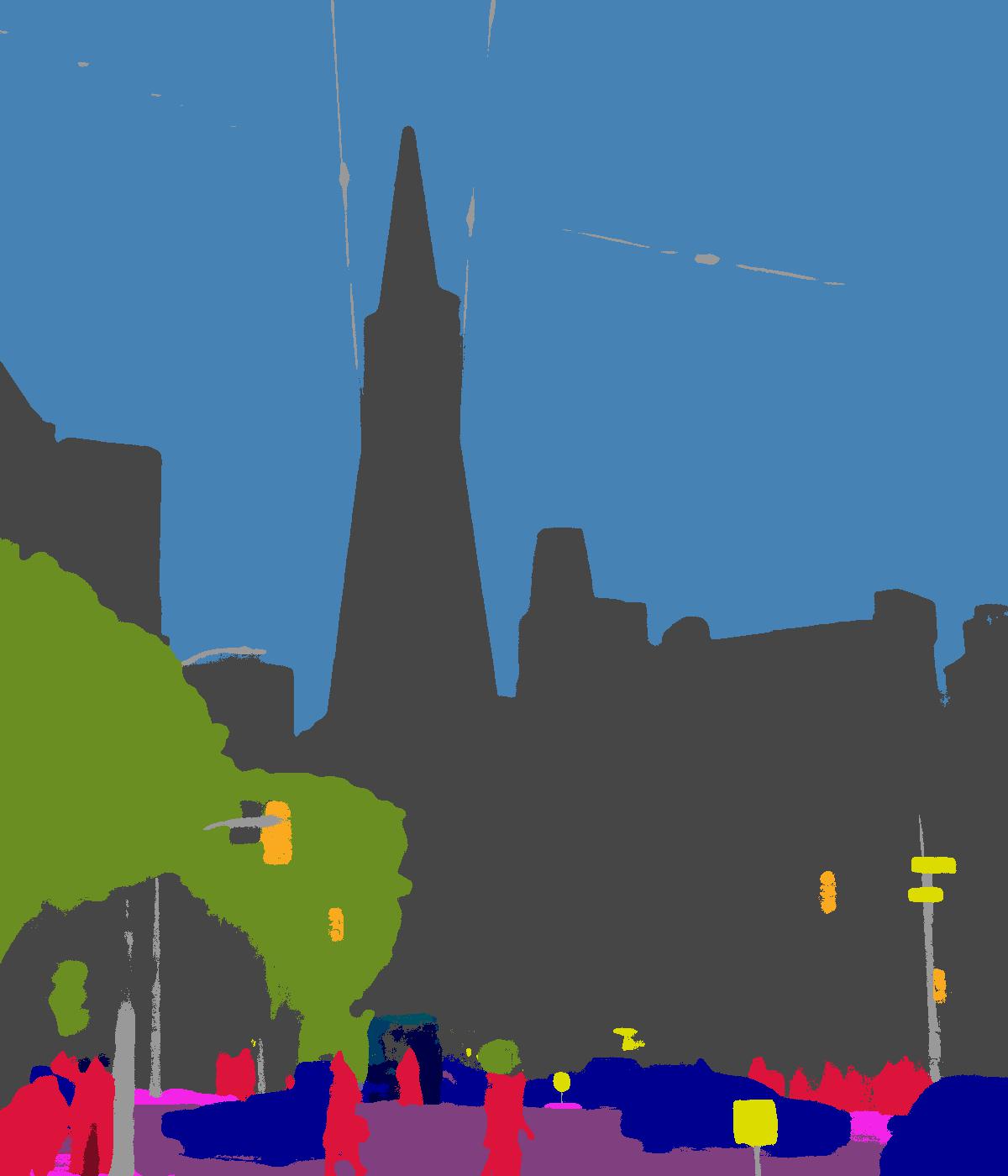} &
        \includegraphics[width=0.165\textwidth,keepaspectratio]{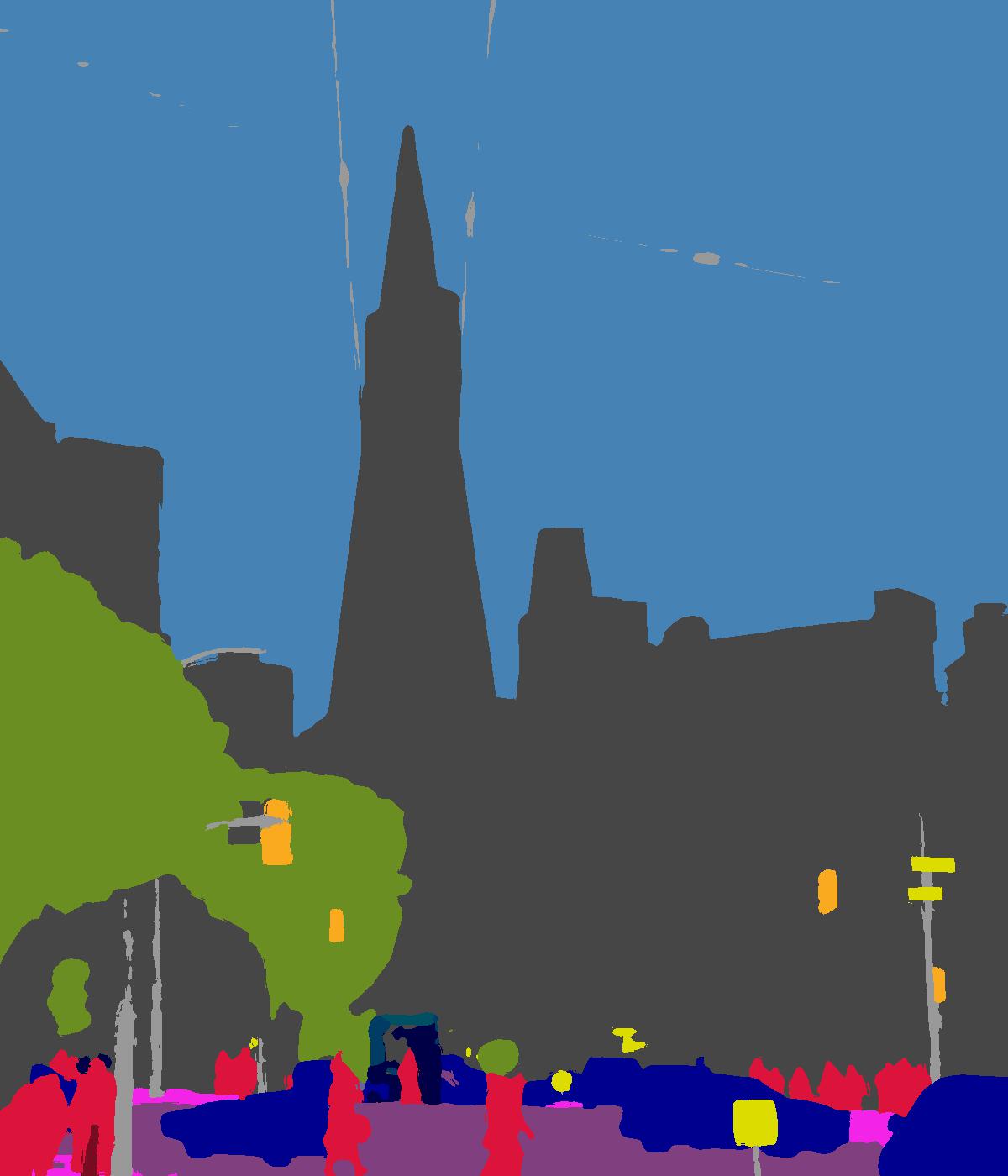}&
        \includegraphics[width=0.165\textwidth,keepaspectratio]{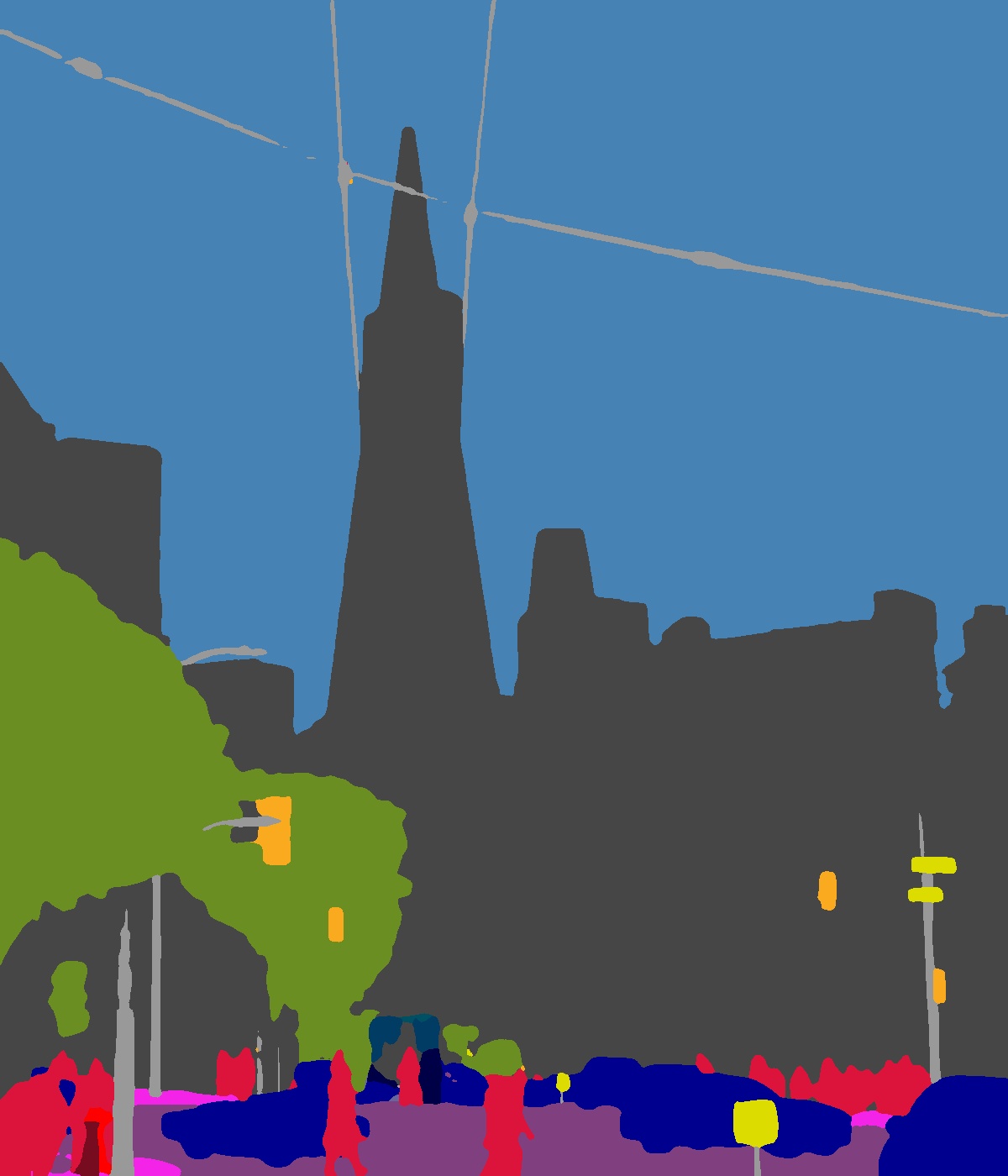}  
        \vspace{-0.065cm}\\
        \includegraphics[width=0.165\textwidth,keepaspectratio]{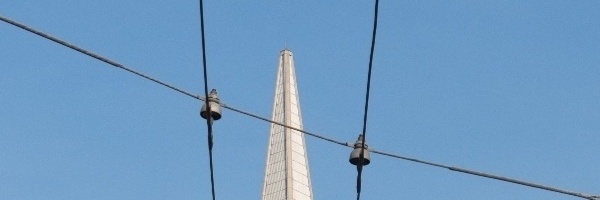} &
        \includegraphics[width=0.165\textwidth,keepaspectratio]{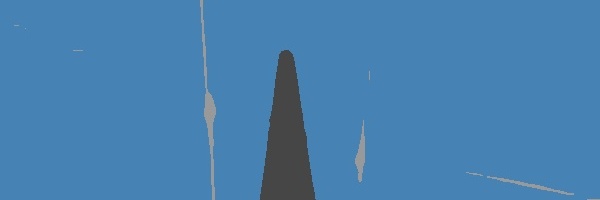} &
        \includegraphics[width=0.165\textwidth,keepaspectratio]{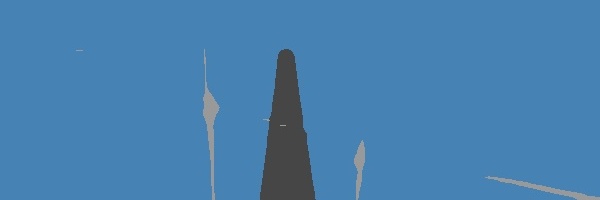} &
        \includegraphics[width=0.165\textwidth,keepaspectratio]{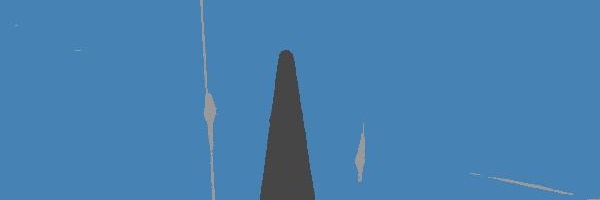} &
        \includegraphics[width=0.165\textwidth,keepaspectratio]{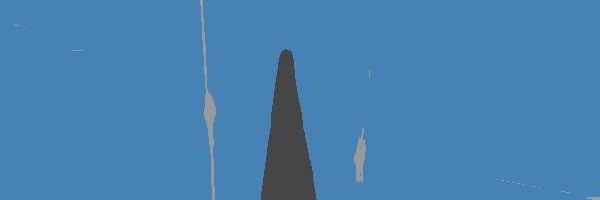} &
        \includegraphics[width=0.165\textwidth,keepaspectratio]{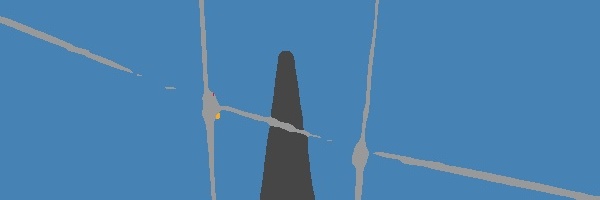}
        \vspace{-0.1cm}\\
        \footnotesize RGB & \footnotesize CE & \footnotesize CIBL & \footnotesize DenseCRF & \footnotesize Segfix & \footnotesize Ours
        \vspace{-0.4cm}\\
        \end{tabular}
		\label{fig:teaser}
        \begin{minipage}[t]{\textwidth} 
		\caption{\small{Segmentation results of an internet image. \tipadd{CE: Produced by DeeplabV3 \protect\cite{chen2017rethinking} trained with cross-entropy loss on Cityscapes \protect\cite{cordts2016cityscapes} dataset. CIBL: Produced by DeeplabV3 trained with cross-entry loss + lov{\'a}sz-softmax \protect\cite{berman2018lovasz} + Boundary Loss \protect\cite{kervadec2019boundary} on Cityscapes dataset. DenseCRF: Refined results of column `CE' by DenseCRF \protect\cite{krahenbuhl2011efficient}. Segfix: Refined results of column `CE' by Segfix \protect\cite{yuan2020segfix}. Ours: Re-trained by adding our loss.}}}
		\label{fig:teaser}
		\end{minipage}
    \vspace{-0.6cm}
\end{figure}
}]

{
  \renewcommand{\thefootnote}{\fnsymbol{footnote}}
  \footnotetext[1]{Corresponding author.}
  \insert\footins{\noindent\footnotesize{Copyright \copyright\space 2022,
Association for the Advancement of Artificial Intelligence (www.aaai.org).
All rights reserved.}} 
}

\begin{abstract}
This paper proposes a novel active boundary loss for semantic segmentation. It can progressively encourage the alignment between predicted boundaries and ground-truth boundaries during end-to-end training, which is not explicitly enforced in commonly used cross-entropy loss. Based on the predicted boundaries detected from the segmentation results using current network parameters, we formulate the boundary alignment problem as a differentiable direction vector prediction problem to guide the movement of predicted boundaries in each iteration. Our loss is model-agnostic and can be plugged in to the training of segmentation networks to improve the boundary details. Experimental results show that training with the active boundary loss can effectively improve the boundary F-score and mean Intersection-over-Union on challenging image and video object segmentation datasets. 
\end{abstract}

\section{Introduction}
Semantic segmentation is a fine-grained, pixel-wise classification task that assigns each pixel a semantic class label to facilitate high-level image analysis and processing. Recently, the accuracy of semantic segmentation has been substantially improved with the introduction of fully convolutional networks (FCNs) \cite{long2015fully, minaee2021image}. \aaaidel{Various forms of multi-scale feature fusion \cite{chen2018encoder,ding2018context,pang2019towards,ronneberger2015u,WangSCJDZLMTWLX19,zhao2017pyramid} and 
non-local attention modules \cite{fu2019adaptive,huang2019ccnet,li2019expectation,zhang2018context,zhao2018psanet} are proposed to further boost the performance of FCNs on large-scale semantic segmentation datasets, such as PASCAL VOC 2012 \cite{everingham2015pascal}, Cityscapes \cite{cordts2016cityscapes}, and ADE20K \cite{zhou2017scene}.}
FCNs leverage convolutional layers and downsampling operations to achieve a large receptive field. Although these operations can encode context information surrounding a pixel, they tend to propagate feature information throughout the image, leading to undesirable feature smoothing across object boundaries. Thus, the segmentation results might be blurred and lack fine object boundary details. To address this issue, boundary-aware information flow control and multi-task training methods have been proposed to improve the discriminative power of features belonging to different objects \cite{bertasius2016semantic,takikawa2019gated,zhu2019improving}. Alternatively, the segmentation errors at boundaries can be remedied by learning the correspondence between a boundary pixel and its corresponding interior pixel \cite{yuan2020segfix}. \aaaidel{The correspondence is represented as a direction vector, and an auxiliary network is trained to extract the features for the purpose of direction prediction.} Despite the empirical success of boundary-aware methods in improving the segmentation accuracy, there still exist a significant amount of segmentation errors at object boundaries, especially for small and thin objects. The mutual dependence between semantic segmentation and boundary detection should be further studied to improve the quality of segmentation results.

In this paper, we propose a novel active boundary loss (ABL) to progressively encourage the alignment between predicted boundaries (PDBs) and ground-truth boundaries (GTBs) during end-to-end training, in which the PDBs are semantic boundaries detected in the segmentation results of the current network. To facilitate end-to-end training, the loss is formulated as a differentiable direction vector prediction problem. Specifically, for a pixel  on the PDBs, we first determine a direction pointing to the closest GTB pixel, and then move the PDB at this pixel towards the direction in a probabilistic manner. \aaaidel{However, there may exist conflicts of direction vectors among adjacent pixels due to the complicated boundary shapes. We thus propose to detach the gradient flow to suppress such conflicts, a detachment which is critical to the success of ABL.} \aaaiadd{Moreover, we also propose to detach the gradient flow to suppress possible conflicts.} Overall, the behavior of ABL is dynamic because the PDBs are changing with the updated network parameters during training. It can be viewed as a variant of classical active contour methods \cite{Kass1988}, since our method first determines the direction vectors in accordance with the PDBs in the current iteration and lets the PDBs move along the direction vectors to reach the GTBs.


\crc{Unlike the cross-entropy loss that only supervises pixel-level classification accuracy, ABL supervises the relationship between PDB and GTB pixels. It embeds boundary information such that the network can pay attention to boundary pixels to improve the segmentation results. 
Moreover, Intersection-over-Union~(IoU) loss pays more attention to overall regions of semantic classes but does not focus on the boundary matching.
Thus, ABL can provide complementary information during the training of the network.} \tipadd{As a result, it can be combined with other loss terms to further improve semantic boundary quality.} In our work, we let the ABL work with the most commonly used cross-entropy loss and the lov{\'a}sz-softmax loss \cite{berman2018lovasz}, a surrogate IoU loss, to significantly improve the boundary details in image segmentation. The lov{\'a}sz-softmax loss is introduced to regularize the training so that the ABL can be used even when the PDBs might be noisy and far from the GTBs\aaaidel{(see Fig.~\ref{fig:a2w_val} for the contribution of each loss)}. The advantage of ABL is that it is model-agnostic and can be plugged into the training of image segmentation networks to improve the boundary details. As illustrated in Fig.~\ref{fig:teaser}, it is beneficial to preserve the boundaries of thin objects that contain a small number of interior pixels.

We tested the ABL with state-of-the-art image segmentation networks, including \aaaiadd{CNN-based networks DeepLabV3 \cite{chen2017rethinking}, OCR network \cite{yuan2019object} and Transformer-based network SwinTransformer \cite{DBLP:journals/corr/abs-2103-14030}. }\aaaidel{DeepLabV3 cite{chen2017rethinking} and OCR network cite{yuan2019object}.} We have also tested the ABL with STM \cite{oh2019video}, a video object segmentation (VOS) network, to show that our loss can be applied to improve VOS results as well. The forward inference stage of these networks remains the same during testing. The experimental results show that training with the ABL can effectively improve the boundary F-score and mean Intersection-over-Union~(mIoU) on challenging segmentation datasets.

\section{Related Work}

\begin{figure*}[t]
    \centering
    \vspace{-1.0em}
    \resizebox{0.83\linewidth}{!}{
     	\includegraphics[width=\textwidth]{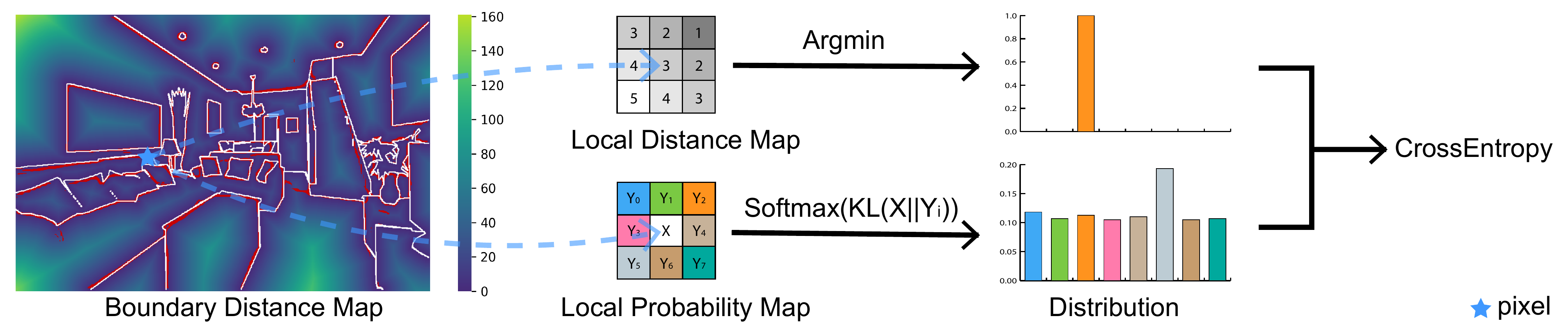}
    }
    \vspace{-5pt}
    \caption{\small{Pipeline of ABL. Boundary distance map is obtained via the distance transform of GTBs, taken an image in the ADE20K~\protect\cite{zhou2017scene} dataset as an example. The overlayed white and red lines in the boundary distance map indicate the GTBs and PDBs, respectively. Local distance map: the number indicates the closest distance to the GTBs. Local probability map: $\mathbf{X}$ and $\mathbf{Y}_i, i \in \{0,1,...,7\}$ denote the class probability distribution for these pixels.}}
    \label{fig:flowchart}
    \vspace{-10pt}
\end{figure*}

\noindent \textbf{FCN-based semantic segmentation.}
FCNs \cite{long2015fully} for semantic segmentation frequently utilize encoder-decoder structure to generate pixel-wise labelling results for high-resolution images. 
\aaaiadd{Successor methods \cite{ronneberger2015u, ding2018context, minaee2021image} are dedicated to a better fusion of multi-scale features to enhance the accuracy of localization and handle small objects.}
FCN-based methods have also been widely used in VOS, including propagation-based methods \cite{hu2017maskrnn, oh2019video, voigtlaender2019feelvos} and detection-based methods \cite{caelles2017one, li2017fully, shin2017pixel}. The key challenge is how to leverage temporal coherence and learn discriminative features of target objects to handle occlusion, appearance change, and fast motion. Since our loss is model-agnostic, \aaaiadd{it can also be applied to VOS for the purpose of boundary refinement.} \aaaidel{we can also apply our loss to VOS for the purpose of boundary refinement.}

\noindent \textbf{Boundary-aware semantic segmentation.} One way to exploit boundary information in deep learning-based semantic segmentation is through multi-task training, in which additional branches are often inserted to detect semantic boundaries \cite{chensupervised,gong2018instance,ruan2019devil,su2019selectivity, xu2018pad, takikawa2019gated, zhu2021unified}. A key challenge in these methods is how to efficiently fuse features from a boundary detection branch to improve semantic segmentation. \aaaidel{GSCNN cite{takikawa2019gated} designs a two-stream network to integrate the shape information for enhancing the performance of semantic segmentation.\citet{li2020improving} proposed a feature-warping module to decouple features from the backbone into the body and boundary features. This module enables the network to supervise body and boundary parts independently in order to reduce feature noise. } \aaaidel{zhen2020joint proposed an iterative semantic boundary and semantic segmentation fusion strategy and designed a pyramid context module to leverage the context from one task to refine the feature map of another task. PAD-Net utilizes the interaction between multi-modal tasks to simultaneously improve the performance of depth estimation and scene parsing.}


There are also works focusing on the control of information flow through boundaries \cite{bertasius2016semantic, ke2018adaptive, bertasius2017convolutional, ding2019boundary, chen2016semantic}.\aaaidel{such as boundary neural fields~cite{bertasius2016semantic}, affinity ﬁeld~cite{ke2018adaptive}, random walk~cite{bertasius2017convolutional}, and boundary-aware feature propagation~cite{ding2019boundary}.} These methods usually learn pairwise pixel-level affinity to maintain the feature difference for pixels near semantic boundaries, while enhancing the similarity of features for interior pixels simultaneously. \aaaidel{Instead of controlling the information flow implicitly through pixel-level affinity, citet{chen2016semantic} proposed an edge-preserving filtering method, which controls the amount of segmentation result smoothing by using a learned reference edge map.} \aaaiadd{Assuming that boundaries can be correlated through a homography transformation, \citet{borse2021inverseform} proposed a frozen inverse transformation network as a boundary-aware loss for boundary distance measurement.}

The boundary details of the segmentation results can also be improved in post-refinement. DenseCRF \cite{krahenbuhl2011efficient} is often used to refine the segmentation results around boundaries. Segfix \cite{yuan2020segfix} trains a separate network to predict the correspondence between boundary and interior pixels. Thus, labels of interior pixels can be transferred to boundary pixels. \tipadd{Although these methods can efficiently refine most boundaries, they fail to model the relationship of pixels inside thin objects that contain a small number of interior pixels, which may downgrade the quality of slender object boundaries, as shown in Fig.~\ref{fig:teaser}. In contrast, the ABL encourages the alignment of PDBs and GTBs. Our experiment shows that it can handle such boundaries well.}
\aaaidel{For instance segmentation, a learning-based deep snake model is proposed to align an initial contour to the object boundary \cite{PengJPLBZ20}. \citet{kirillov2020pointrend} proposed a multi-resolution refinement method to refine the segmentation result by propagating boundary information from low-resolution levels to high-resolution levels. CascadePSP performs global and local steps to capture high-resolution object boundary details.}

The uniqueness of our ABL is that it allows propagating the GTB information with a distance transform for regulating the network behavior at the PDBs, while the network structure can remain the same. As a loss, ABL can save efforts in network design. \citet{kervadec2019boundary} proposed Boundary Loss~(BL) for image segmentation, which is most related to our work. However, this loss is designed for unbalanced binary segmentation and actually a regional IoU loss. In our implementation, the ABL is coupled with an IoU loss in \cite{berman2018lovasz} to further refine the boundary details.

\section{Active Boundary Loss}

The ABL continuously monitors the changes on the PDBs in segmentation results to determine the plausible moving directions. Its computation is divided into two phases. First, for each pixel $i$ on the PDBs, we determine its next candidate boundary pixel $j$ closer to the GTBs in accordance with the relative location between the PDBs and GTBs. Second, we use the KL divergence as logits to encourage the increase in KL divergence between the class probability distribution of $i$ and $j$. Meanwhile, this process reduces the KL divergence between $i$ and the rest of its neighboring pixels. In this way, the PDBs can be gradually pushed towards the GTBs. Unfortunately, candidate boundary pixel conflicts might occur, severely degrading the performance of the ABL. Thus, we carefully reduce the conflicts through gradient flow control in the computation of ABL, which is crucial to its success. 

The overall pipeline of ABL is illustrated in Fig.~\ref{fig:flowchart}. Each phase and how to suppress the conflicts are detailed as follows. Hereafter, we use $\mathbf{A}_i$ to denote the value stored at pixel $i$ of a map $\mathbf{A}$.

\noindent \textbf{Phase I.} This phase starts with detecting the PDBs using the class probability map $\mathbf{P} \in \mathbb{R}^{C \times H \times W}$ output by the current network, where $C$ denotes the number of semantic classes, and the image resolution is $H \times W$. Specifically, we compute a boundary map $\mathbf{B}$ through the computation of KL-divergence to indicate the locations of PDBs. For a pixel $i$ in $\mathbf{B}$, its value $\mathbf{B}_i$ is computed as follows:  
\vspace{-2pt}
\begin{equation}
    \mathbf{B}_{i}= \Big\{
    \begin{array}{cl}
        1 &\text{if } \exists \mathbf{KL}(\mathbf{P}_{i},\mathbf{P}_{j}) > \epsilon, j \in \mathcal{N}_2(i)\,; \\
        0 &Otherwise  \,,
    \end{array} 
\label{eq:BoundSoftPred}
\vspace{-2pt}
\end{equation}
where $1$ indicates the existence of PDBs, and $\mathbf{P}_i$ is the $C$-dimensional vector extracted from the probability map at pixel $i$. $\mathcal{N}_2(i)$ indicates the 2-neighborhood of pixel $i$. Specifically, the offsets of pixels in $\mathcal{N}_2(i)$ to pixel $i$ are  $\{\{1, 0\},\{0, 1\}\}$. \tipadd{Since it's difficult to define a perfect fixed threshold $\epsilon$ to detect PDBs, we choose an adaptive threshold to ensure that the number of boundary pixels in $\mathbf{B}$ is less than $1\%$ of the total pixels of the input image, where $1\%$ is a ratio to approximate the number of boundary pixels in an image.}
Empirically, we observe that setting $\epsilon$ in this adaptive way can largely avoid the emergence of excessive misleading pixels in $\mathbf{B}$ far from the GTBs, especially in the early training period.
\tipadd{Controlling boundary pixel number also helps to save the computational cost of ABL.}

\crc{Subsequently, for a pixel $i$ on PDBs, its next candidate boundary pixel $j$ is selected as its neighboring pixel with the smallest distance value computed by the distance transform\footnotemark[1] of the GTBs.} 
The GTBs are also determined using Eq.~\ref{eq:BoundSoftPred}, but the KL divergence is replaced by checking whether the ground-truth class labels are equal between pixel $i$ and $j \in \mathbf{N}_2(i)$. To represent the coordinate of pixel $j$ in the computation of ABL, we convert it into an offset to pixel $i$ and then encode it as a one-hot vector. Specifically, we compute a target direction map $\mathbf{D}^g \in \{ 0, 1 \}^{8 \times H\times W}$, where the one-hot vector for a pixel $i$ stored at $\mathbf{D}^g_i$ is 8D, because we use 8-neighborhood in this operation. The formula to compute $\mathbf{D}^g_i$ can be written as:
\vspace{-1pt}
\begin{equation}
    \mathbf{D}^g_i = \Phi({\arg\min}_j \ \mathbf{M}_{i+{\Delta}_j})\ , \ \  j \in \{0,1,...,7\},
\label{eq:DirectionGT}
\vspace{-2pt}
\end{equation}
where $\Delta_j$ represents the $j^{\text{th}}$ element in the set of directions $\Delta=$ $\{\{1,0\},$ $\{-1,0\},$ $\{0,-1\},$ $\{0,1\},$ $\{-1,1\},$ $\{1,1\},$ $\{-1,-1\},$ $\{1,-1\}\}$, and $\mathbf{M}$ is the result of distance transform of GTBs. The pixel $i+{\Delta}_j$ with the smallest distance is selected as the next candidate boundary pixel. The function $\Phi$ converts index $j$ into a one-hot vector. For instance, if $j=1$, $\Phi(j)$ should be $\{0,1,0,0,0,0,0,0\}$, which is similar to the direction representation used in Segfix. In implementation, we dilate $\mathbf{B}$ with $1$ pixel and perform this operation for all the pixels in dilated $\mathbf{B}$ to accelerate the movement of the PDBs, since more pixels are covered. 

\footnotetext[1]{$scipy.ndimage.morphology.distance\_transform\_edt$ is used in the implementation of distance transform.}

\noindent \textbf{Phase II.} The 8D vector $\mathbf{D}^g_i$ computed in Eq.~\ref{eq:DirectionGT} is set to be the target distribution in the cross-entropy loss. We aim to increase the KL divergence between the class probability distribution of $i$ and $j$, and simultaneously reduce the KL divergence between $i$ and the rest of its neighboring pixels. An 8D vector using the KL divergence between pixel $i$ and its neighboring pixel $j$ as logits, denoted by $\mathbf{D}^{p}_{i}$, is then computed as follows:
 \vspace{-1pt}
\begin{equation}
\mathbf{D}^{p}_{i} = \left\{\frac{{\rm e}^{{\mathbf{KL}({\mathbf{P}_{i}}}, {\mathbf{P}_{{i+\mathbf{\Delta}_k}}})}}{\sum_{m=0}^{7}{\rm e}^ {\mathbf{KL}({\mathbf{P}_{i}}, {\mathbf{P}_{{i+\mathbf{\Delta}_m}}})}}, k \in \{0,1,...,7\} \right\},
\label{eq:DirectionPred}
 \vspace{-1pt}
\end{equation}
where $\mathbf{KL}$ indicates the function to compute the KL divergence using $\mathbf{P}_{i}$ and $\mathbf{P}_{{i+\mathbf{\Delta}_k}}$.

For those pixels on the PDBs, the ABL is computed as the weighted cross-entropy loss:
\vspace{-4pt}
\begin{equation}
\mathbf{ABL} = \frac{1}{N_b}\sum_i^{N_b} \mathbf{\Lambda}(\mathbf{M}_i) \mathbf{CE}\left(\mathbf{D}^{p}_{i}, \mathbf{D}^{g}_{i} \right) .
\label{eq:LossEdgeSimple}
\vspace{-4pt}
\end{equation}
The weight function $\mathbf{\Lambda}$ is computed as $\mathbf{\Lambda}(x) = \frac{\mathbf{min}(x, \theta)}{\theta}$,
where \blue{$N_b$ is the number of pixels on the PDBs} and $\theta$ is a hyper-parameter set to 20. The closest distance to the GTBs at pixel $i$ is used as a weight to penalize its deviation from the GTBs. If $\mathbf{M}_{i}$ is $0$, indicating that the pixel is already on the GTBs, this pixel will be discarded in the ABL.

\begin{figure}[t!]
\vspace{-1em}
\begin{center}
    \setlength{\tabcolsep}{1pt}
    \resizebox{\linewidth}{!}{
        \includegraphics[]{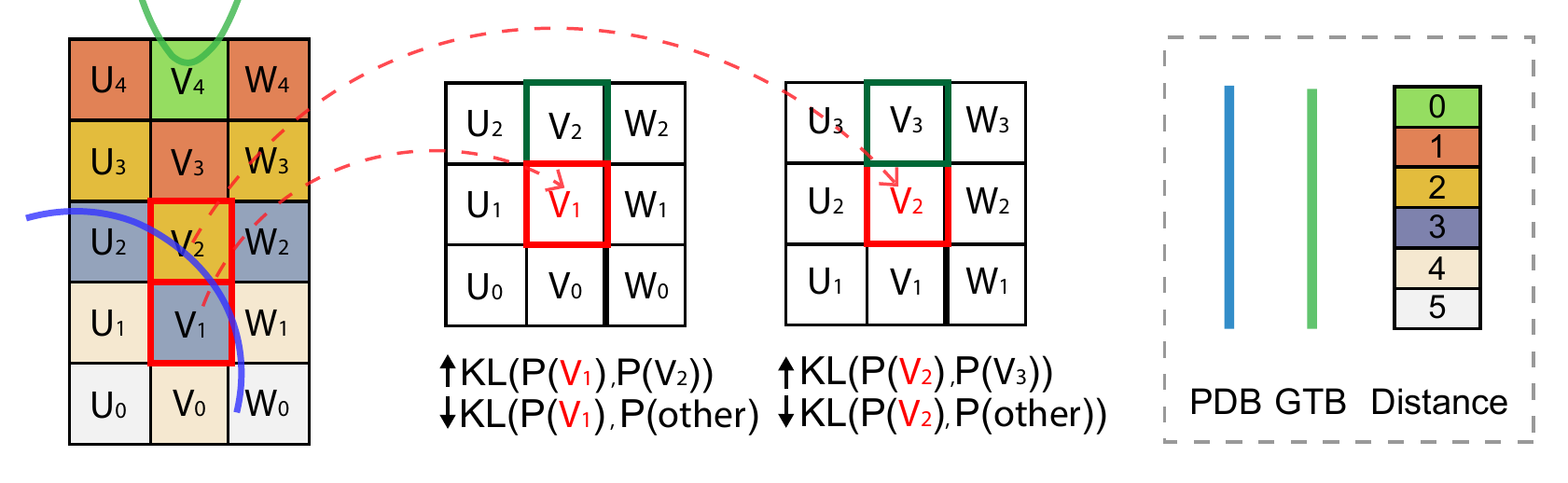}
    }
\end{center}
\vspace{-10pt}
\caption{An example of conflict. $V_4$: pixel on a GTB. $\uparrow$: increase. $\downarrow$: decrease. The KL divergence between $V_1$ and $V_2$ is required to increase for $V_1$ but to decrease for $V_2$, resulting in contradictory gradients for $V_1$ and $V_2$. } 
\label{fig:confusion}
\vspace{-10pt}
\end{figure}

\noindent\textbf{Conflict suppression.} Determining pixels on the PDBs using KL divergence might lead to the conflict case, as shown in Fig.~\ref{fig:confusion}. In this case, pixels $V_1$ and $V_2$ are deemed to be on a PDB (indicated by the red curve) because the KL divergence values computed for ($V_1, W_1$) and ($V_2, V_3$) are larger than the threshold. However, the GTB (indicated by the green curve) leads to the conflict when computing the ABL for $V_1$ and $V_2$ because the GTB is to the right of $V_1$ and $V_2$. Thus, for pixel $V_1$, we need to \emph{increase} $\mathbf{KL}(\mathbf{P}_{V_1}, \mathbf{P}_{V_2})$ because $V_2$ is the closest to the GTB and it should be the next candidate pixel in the neighborhood of $V_1$. In contrast, for pixel $V_2$, we need to \emph{decrease} $\mathbf{KL}(\mathbf{P}_{V_2}, \mathbf{P}_{V_1})$ because pixel $V_3$ is the next candidate boundary pixel for $V_2$ rather than $V_1$. Thus, the gradients of the ABL computed for $\mathbf{P}_{V_1}$ and $\mathbf{P}_{V_2}$ might contradict with each other. 

\blue{While it might be possible to design a global search algorithm to remove such kind of conflicts, it will significantly slow down the training.} Thus, we choose to suppress the conflicts through the easy-to-implement detaching operation in Pytorch. Specifically, through the detaching operation, the gradient of ABL is computed only for the pixels on the PDBs, but not for its neighboring pixels. This process indicates that, for a $3 \times 3$ patch, we focus on the adjustment of the class probability distribution of pixels on the PDBs only so as to move the PDBs towards the GTBs. As a result, the conflicting gradient flow from $\mathbf{KL}(\mathbf{P}_{V_2}, \mathbf{P}_{V_1})$ to $\mathbf{P}_{V_1}$ is blocked in this case, and vice versa. Empirically, we found the mIoU drops around $3\%$ without the detaching operation. 

Furthermore, we use label smoothing \cite{szegedy2016rethinking} to regularize the ABL by setting the largest probability of the one-hot target probability distribution to $0.8$ and the rest to $0.2/7$ (the parameters, $0.8$ and $0.2/7$ are determined through experiments). This process can avoid over-confident decisions of network parameter updating, especially when there exist several pixels with the same distance value in the neighborhood of pixels on the PDBs. The detaching operation is also beneficial in this case to avoid conflicts in the gradient flow. 

\section{Training Loss}

The training loss $\mathcal{L}_t$ we mainly used to train a semantic segmentation network consists of three terms: 
\vspace{-3pt}
\begin{equation}
    \mathcal{L}_t =  \mathbf{CE} + \mathbf{IoU} + w_{a}\mathbf{ABL} ,
\label{eq:LossTrain}
\vspace{-3pt}
\end{equation}
where $\mathbf{CE}$ is the most commonly used cross-entropy (CE) loss, which focuses on the per-pixel classification. The combination of lov{\'a}sz-softmax loss, namely $\mathbf{IoU}$, and our ABL are two loss terms that are added to improve the boundary details, and $w_{a}$ is a weight. The lov{\'a}sz-softmax loss is expressed as follows \cite{berman2018lovasz}: 
\vspace{-3pt}
\begin{equation}
    \mathbf{IoU}=\frac{1}{|\mathcal{C}|} \sum_{c \in \mathcal{C}} \overline{\Delta_{J_{c}}}(\mathbf{m}(c)) ,
\label{eq:LossIoU}
\vspace{-3pt}
\end{equation}
where $\mathcal{C}$ is the number of classes, and  $\mathbf{m}(c)$ is the vector of prediction errors for class $c \in \mathcal{C}$. $\overline{\Delta_{J_{c}}}$ indicates the lov{\'a}sz extension of the Jaccard loss $\Delta_{J_{c}}$. 

The reason for introducing the lov{\'a}sz-softmax loss is twofold: 1)  This loss tends to prevent small objects from being ignored in segmentation such that the $\mathbf{ABL}$ can be used to improve their boundary details, since the $\mathbf{ABL}$ relies on the existence of predicted boundaries as the beginning step of its computation. 2) It can balance with the noisy predicted boundary pixels, especially at the early training period. The improvement of $\mathbf{ABL}$ over $\mathbf{CE}$ plus $\mathbf{IoU}$ is verified in the \aaaiadd{Experiments section}. \aaaidel{(Sec.~\ref{sec:experiments}).}


\section{Experiments}
\label{sec:experiments}

We implemented the ABL on a GPU server (2 Intel Xeon Gold 6148 CPUs, 512GB memory) with 4 Nvidia Tesla V100 GPUs. In this section, we report ablation studies, quantitative and qualitative results obtained from the evaluation of the ABL in image segmentation experiments and a test of fine-tuning VOS network. 

\noindent \textbf{Baselines.} We use the OCR network \cite{yuan2019object} [backbone: HRNetV2-W48 \cite{WangSCJDZLMTWLX19}], DeeplabV3 \cite{chen2017rethinking} [backbone: ResNet-50 \cite{he2016deep}], \aaaiadd{and UperNet \cite{xiao2018unified} [backbone: SwinTransformer \cite{DBLP:journals/corr/abs-2103-14030}]} as the baseline models for the task of semantic image segmentation. To verify that our ABL can be applied to the task of video object segmentation, we use STM \cite{oh2019video} as the baseline, since its pre-trained model is publicly available.

\noindent \textbf{Dataset.} We evaluate our loss mainly on the image segmentation dataset Cityscapes \cite{cordts2016cityscapes} and ADE20K \cite{zhou2017scene}. These two datasets provide densely annotated images that are important for the training of our method to align semantic boundaries. Cityscapes dataset contains high-quality dense annotations of 5000 images with 19 object classes, and ADE20K is a more challenging dataset with 150 object classes. There are 20210/2000/3000 images for the training/validation/testing set in ADE20K, respectively. 
Following the training protocol of \cite{yuan2019object}, we use random crop, scaling (from $0.5$ to $2$), left-right flipping and brightness jittering between $-10$ and $10$ degrees in data augmentation. In multi-scale inference, we apply scales $\{0.5,0.75,1.0,1.25,1.5,1.75,2.0\}$ and $\{0.5,0.75,1.0,1.25,1.5,1.75\}$ as well as their mirrors. \\
\noindent \textbf{Training parameters.} We use stochastic gradient descent as the optimizer and utilize a ``ploy'' learning rate policy similar to \citet{chen2017deeplab} in the training. Hence, the initial learning rate is multiplied by $(1-\frac{iter}{max iter})^{power}$ with power = $0.9$. Sync Batch Normalization \cite{zhang2018context} is used in all our experiments to improve stability. The detailed training and testing settings for ADE20K and Cityscapes are as follows: \\
$\bullet$ ADE20K: the parameters are set as follows: initial learning rate = $0.02$, weight decay = $0.0001$, crop size = $520\times 520$, batch size = $16$, and 150k training iterations, which are the same as the setting in \citet{yuan2019object}.\\ 
$\bullet$ Cityscapes: the parameters are set as follows: initial learning rate = $0.01$ or $0.04$, crop size = $512\times 1024$ (used in OCR) or $769\times 769$ (used in DeeplabV3), weight decay = $0.0005$, batch size = $8$, and 80K training iterations. The parameter setting is the same as the setting in \citet{yuan2020segfix}, but we do not use coarse data in experiments. \\

\begin{table}[t]
	\vspace{-2.0em}
	\renewcommand\arraystretch{0.9}
    \small
    \centering
    \resizebox{1.0\linewidth}{!}{
    \begin{tabularx}{1.1\linewidth}{lYYYY}
        \toprule[1pt]
        loss & mIoU & pixAcc \\
        \midrule
        CE & 79.5 & 96.3 \\
        CE + ABL(20\%) & 79.6 & 96.3 \\
        CE + IoU & 80.2 & 96.3 \\
        \tipadd{CE + IoU + BL} & 80.2 & \textbf{96.4} \\
        CE + IABL & \textbf{80.5} & \textbf{96.4} \\
        CE + IABL w/o detach & 76.0 & 95.6 \\ 
        
        \bottomrule[1pt]
    \end{tabularx}
    }
    \vspace{-7pt}
    \caption{The influence of loss terms on Cityscapes dataset. These experiments are conducted using DeeplabV3 network and single-scale inference. ABL(20\%): addin additional ABL in the last 20\% epochs.}
    \label{tab:ablation_detach}
    \vspace{-7pt}
\end{table}

\begin{table}[t]
	\renewcommand\arraystretch{0.9}
    \small
    \centering
    \resizebox{1.0\linewidth}{!}{
    \begin{tabularx}{1.1\linewidth}{T@{\hspace{0.2in}}t@{\hspace{0.07in}}QQQQ}
        \toprule[1pt]
        \multirow{2}{*}{Method} & \multirow{2}{*}{loss} & \multicolumn{2}{c}{mIoU} & \multicolumn{2}{c}{pixAcc} \\ \cmidrule(l){3-4} \cmidrule(l){5-6}
        & & SS & MS & SS & MS \\
        \midrule
        \multirow{4}{*}{\tabincell{l}{OCR~\shortcite{yuan2019object} \\ \thinspace [HRNetV2\\-W48~\shortcite{WangSCJDZLMTWLX19}]}} & CE & 44.51 & 45.66 & \textbf{81.66} & 82.20 \\
        & CE + IoU & 44.73 & 46.54 & 81.52 & 82.29 \\
        & CE + IABL & \textbf{45.38} & \textbf{46.88}& 81.63 & \textbf{82.43}  \\
        & CE + IFKL & 45.11 & 45.96 & 81.61 & 82.28  \\
        \midrule
        \aaaiadd{\multirow{4}{*}{\tabincell{l}{UperNet~\shortcite{xiao2018unified} \\ \thinspace [Swin-T~\shortcite{DBLP:journals/corr/abs-2103-14030}]}}} & CE & 44.51 & 45.81 & 81.09 & 81.96 \\
        & CE~+~IoU & 45.39 & 47.15 & 81.20 & 82.22 \\
        & CE + IABL & \textbf{45.98} & \textbf{47.58} & \textbf{81.34} & \textbf{82.39} \\
        & CE~+~IoU~+~BL & 45.72 & 47.50 & 81.33 & \textbf{82.39} \\
        \bottomrule[1pt]
    \end{tabularx}
    }
    \vspace{-7pt}
    \caption{The influence of loss terms on ADE20K dataset. SS: single-scale inference. MS: multi-scale inference.}
    \label{tab:ablation_cie}
    \vspace{-15pt}
\end{table}

\noindent \textbf{Evaluation Metrics.} Three metrics, \ie pixel accuracy (pixAcc), mean Intersection-over-Union (mIoU), and boundary F-score \cite{perazzi2016benchmark, yuan2020segfix}, are used to demonstrate the performance of the ABL. The first two metrics are used to evaluate the pixel-level and region-level accuracy of a segmentation result, respectively. Boundary F-scores are used to measure the quality of boundary alignment and computed within the area of the dilated GTBs. The dilation parameters are set to $1$, $3$, $5$ pixels in our implementation. To better preserve boundary details in the evaluation, we do not use resize operation in the testing.

%

\noindent\textbf{Combination of loss terms.} To ease the description of the ablation study, we denote different combinations of loss terms used in the training as follows: CE = cross-entropy; CE+IoU = cross-entropy + lov{\'a}sz-softmax; CE+IABL = cross-entropy + lov{\'a}sz-softmax + ABL. $w_{a}$ is set to $1.0$ for ADE20K dataset but $1.5$ for Cityscapes dataset, since training images' resolution is much larger for Cityscapes. 

In addition, we rely on the KL divergence of the class probability distributions of adjacent pixels, which can be viewed as the pair-wise term used in condition random field~\cite{lafferty_conditional_nodate}. Hence, it is necessary to verify how simply enforcing the KL divergence loss at each edge of an image works in the image segmentation, \ie enforcing the loss for each edge between a pair of adjacent pixels, not only at semantic boundaries. To this end, we define a full KL-divergence (FKL) loss as follows:
\begin{equation}
    \resizebox{0.9\linewidth}{!}{
    $\mathbf{FKL} = \frac{1}{N_e}\sum_e\mathbf{BCE}(\frac{1}{1+{\rm e}^{\mathbf{KL}(\mathbf{P}_{e_i}, \mathbf{P}_{e_j})}}, (\mathbf{G}_{e_i} \neq \mathbf{G}_{e_j})),$
    }
\label{eq:FKL}
\end{equation}
where $e$ denotes an image edge that connects a pair of pixels $e_i$ and $e_j$, $N_e$ is the total number of edges in an image, and $(\mathbf{G}_{e_i} \neq \mathbf{G}_{e_j})$ returns $1$ if the ground-truth label of pixel $e_i$ is not equal to the label of $e_j$, otherwise $0$. If the $\mathbf{FKL}$ loss is used with cross-entropy and lov{\'a}sz-softmax loss in the training, we denote this combination as CE+IFKL.



\begin{table}[t]
	\vspace{-2.0em}
	\renewcommand\arraystretch{0.9}
    \small
    \centering
    \resizebox{0.9\linewidth}{!}{
    \begin{tabularx}{\linewidth}{rTvvv}
        \toprule[1pt]
        Method & Backbone & OS & mIoU & pixAcc \\
        \midrule
        CPN~\shortcite{yu2020context} & ResNet-101+CPL & 8$\times$ & 46.27 & 81.85 \\
        PyConv~\shortcite{duta2020pyramidal} & PyConvResNet-101 & 8$\times$ & 44.58 & 81.77 \\
        ACNet\shortcite{fu2019adaptive} & ResNet-101+MG & 4$\times$ & 45.90 & 81.96 \\
        DNL~\shortcite{yin2020disentangled} & HRNetV2-W48 & 4$\times$ & 45.82 & - \\
        \midrule
        OCR~\shortcite{yuan2019object} & HRNetV2-W48 & 4$\times$ & 45.66 & 82.20 \\
        OCR+IABL & HRNetV2-W48 & 4$\times$ & \textbf{46.88} & \textbf{82.43} \\
        \midrule
        \aaaiadd{UperNet}~\shortcite{xiao2018unified} & Swin-B~\shortcite{DBLP:journals/corr/abs-2103-14030} & 4$\times$ & 51.66 & 84.06 \\
        UperNet+IABL & Swin-B & 4$\times$ & \textbf{52.40} & \textbf{84.11} \\
        \bottomrule[1pt]
    \end{tabularx}
    }
    \vspace{-7pt}
    \caption{Results on ADE20K validation set. OS: Output stride. All results are obtained using multi-scale inference.}
    \label{tab:ade20k_big}
    \vspace{-10pt}
\end{table}

\begin{table}[t]
	\renewcommand\arraystretch{0.9}
    \small
    \centering
    \resizebox{0.9\linewidth}{!}{
    \begin{tabularx}{\linewidth}{rlvv}
        \toprule[1pt]
        Method & Backbone & OS & mIoU \\
        \midrule
        GSCNN~\shortcite{takikawa2019gated} & WideResNet-38+ASPP & 8$\times$ & 80.8 \\
        DANet~\shortcite{fu2019dual} & ResNet-101+MG & 8$\times$ & 81.5 \\
        ACNet~\shortcite{fu2019adaptive} & ResNet-101+MG & 4$\times$ & 82.0 \\
        \midrule
        OCR~\shortcite{yuan2019object} & HRNetV2-W48 & 4$\times$ & 82.2 \\ 
        OCR+IABL & HRNetV2-W48 & 4$\times$ & \textbf{82.9} \\
        \bottomrule[1pt]
    \end{tabularx}
    }
    \vspace{-7pt}
    \caption{Results on Cityscapes validation set. OS: Output stride. All results are obtained using multi-scale inference.}
    \label{tab:cityscapes_big}
    \vspace{-15pt}
\end{table}

\begin{figure*}[t]
    \centering
    \vspace{-2.0em}
    \resizebox{0.85\linewidth}{!}{
    	\includegraphics[width=\textwidth]{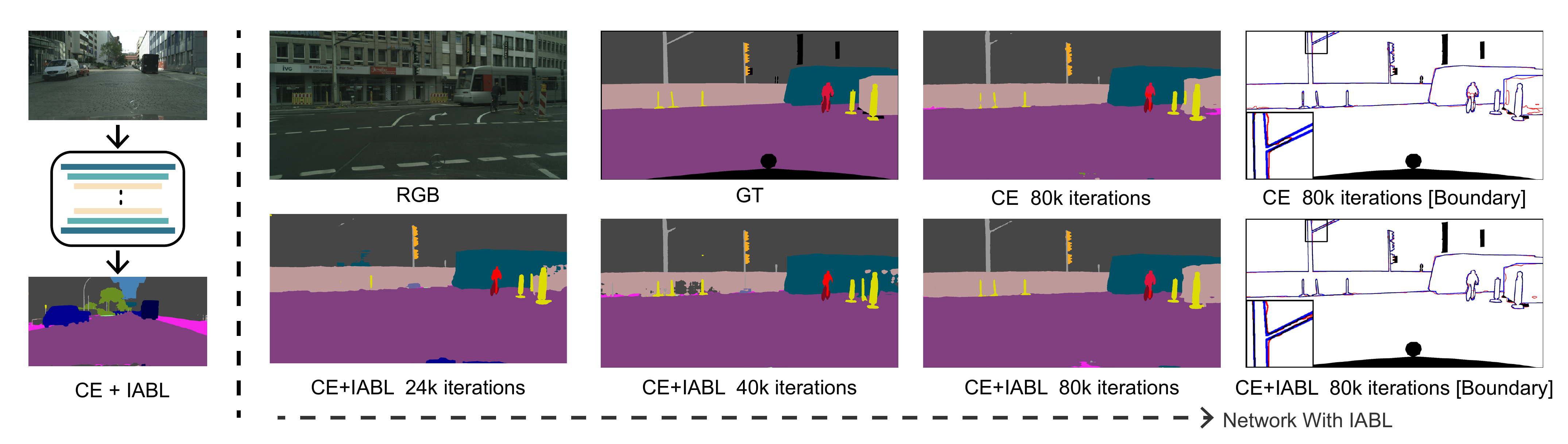}
    }
    \vspace{-10pt}
    \caption{Progressive refinement of boundary details in the training. Dataset: Cityscapes. Network: DeepLabV3. The input image is taken from the Cityscapes training set as an example. The GTBs are in blue and PDBs in red.}
    \label{fig:compare_CE}
    \vspace{-15pt}
\end{figure*}

\begin{figure}[t]
\begin{center}
    \setlength{\tabcolsep}{1pt}
    \resizebox{0.85\linewidth}{!}{
\begin{tabular}{lccccc}
        \multirow{2}{*}{\rotatebox{90}{Cityscapes}} &
        \includegraphics[width=0.15\textwidth]{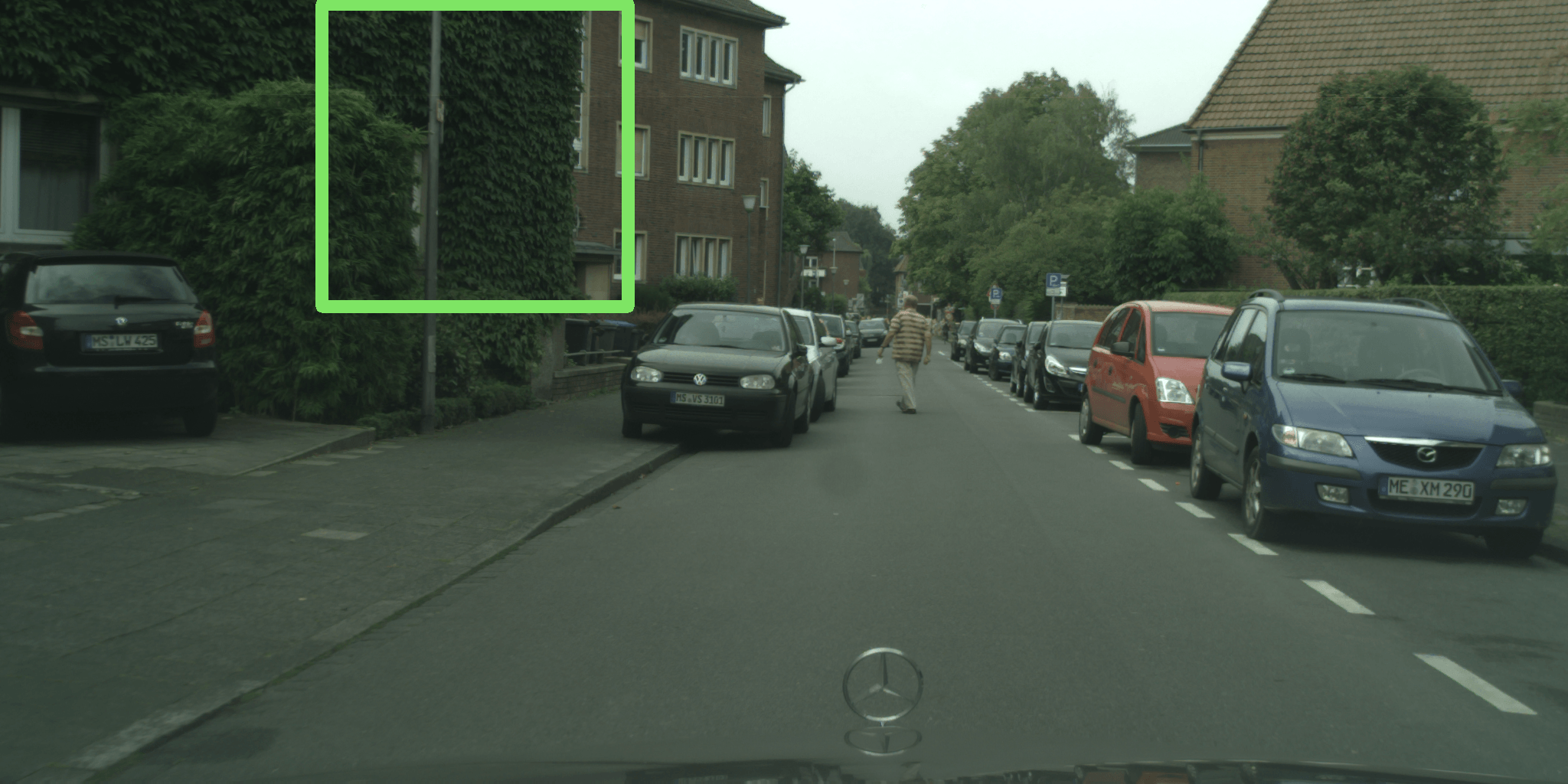} & 
        \includegraphics[width=0.15\textwidth]{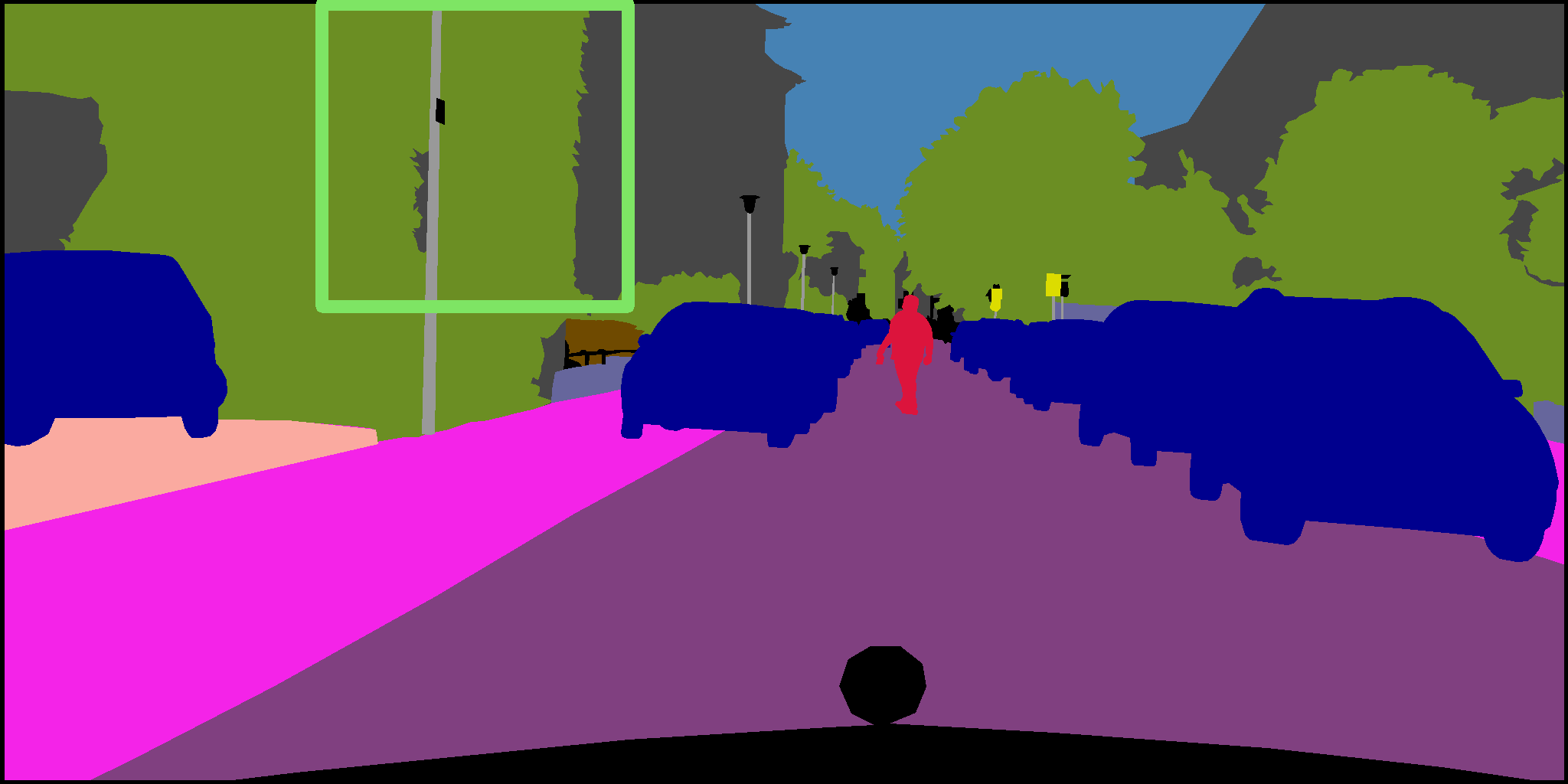} &
        \includegraphics[width=0.15\textwidth]{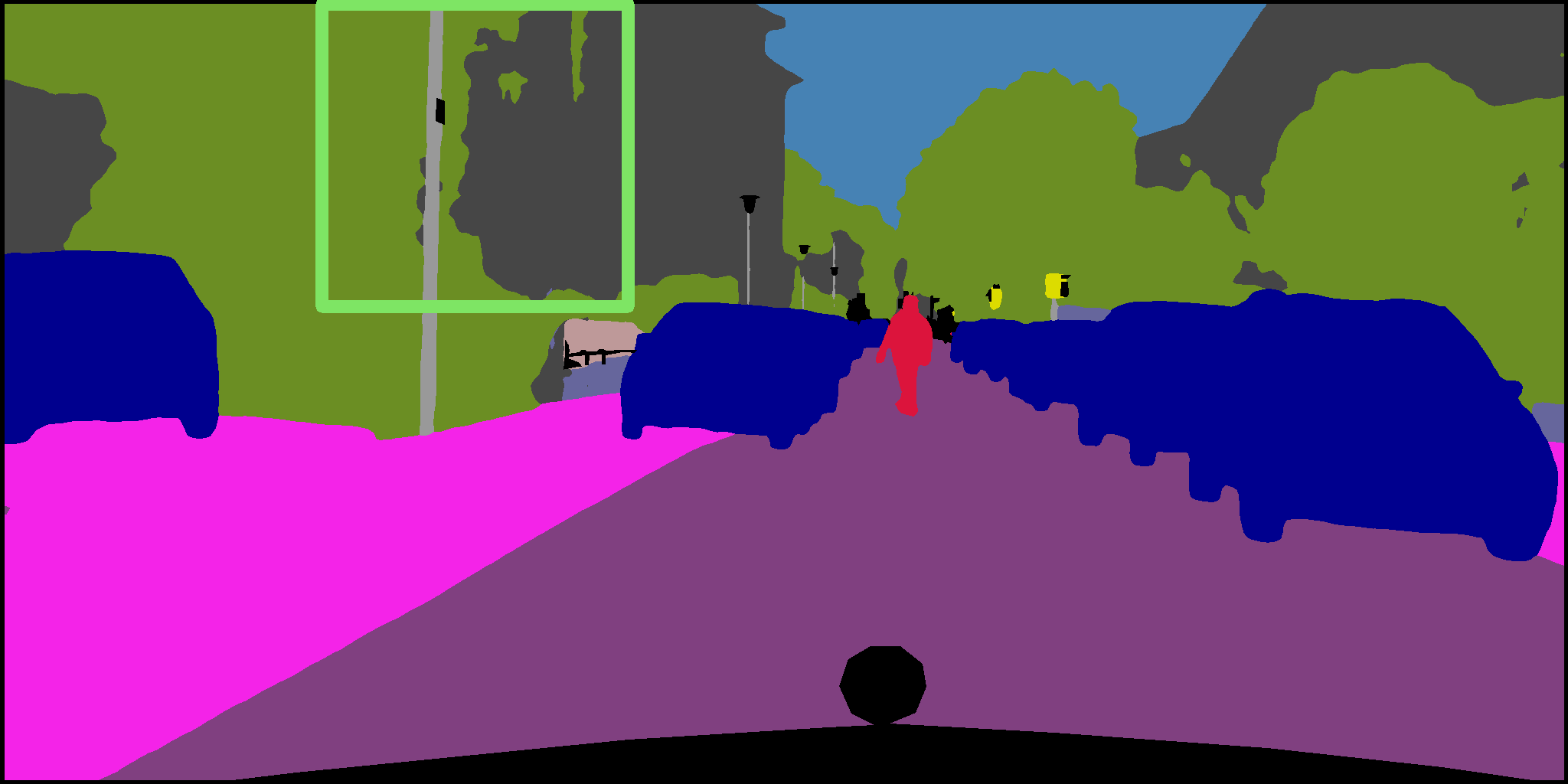} &
        \includegraphics[width=0.15\textwidth]{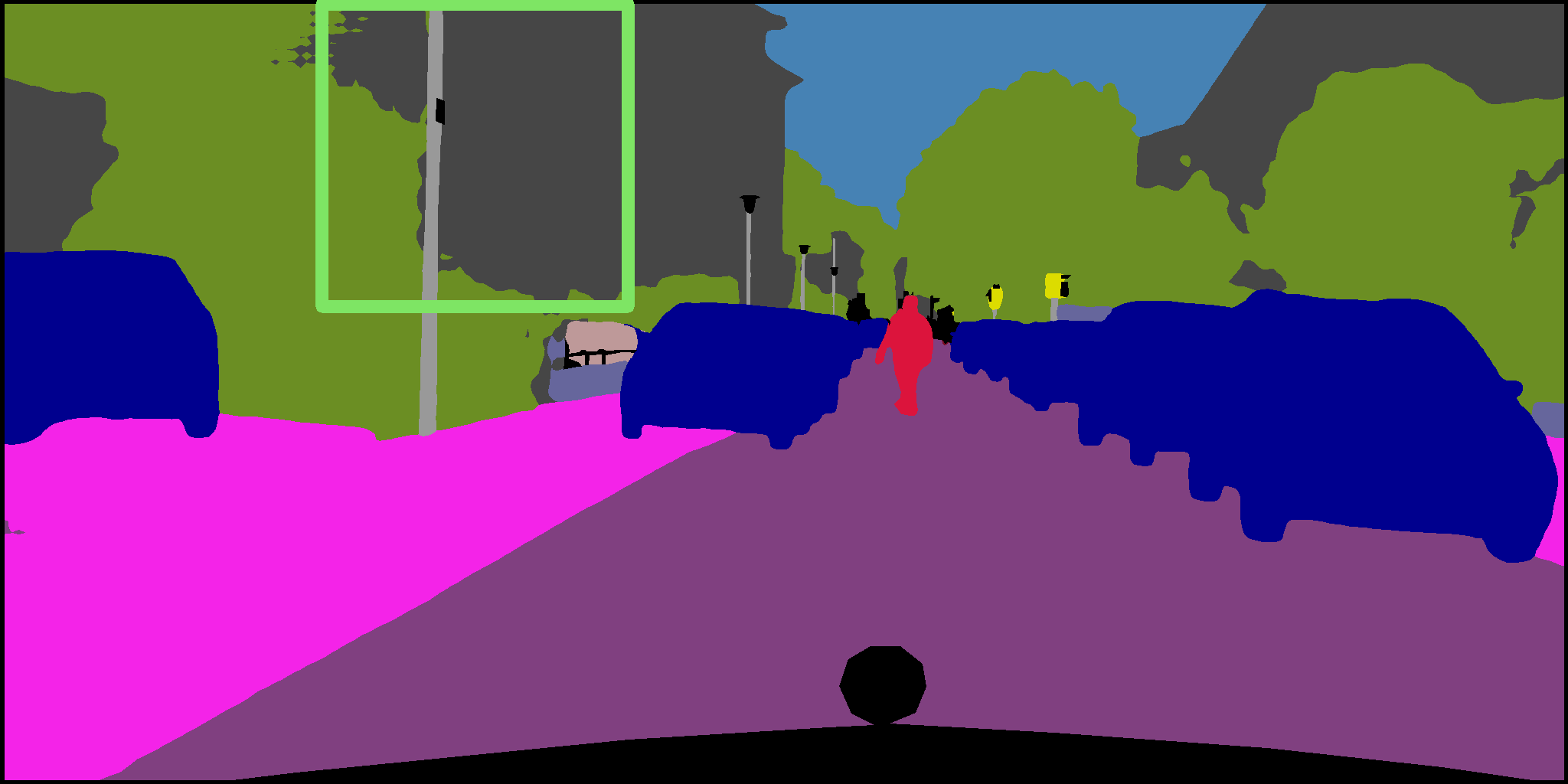} &
        \includegraphics[width=0.15\textwidth]{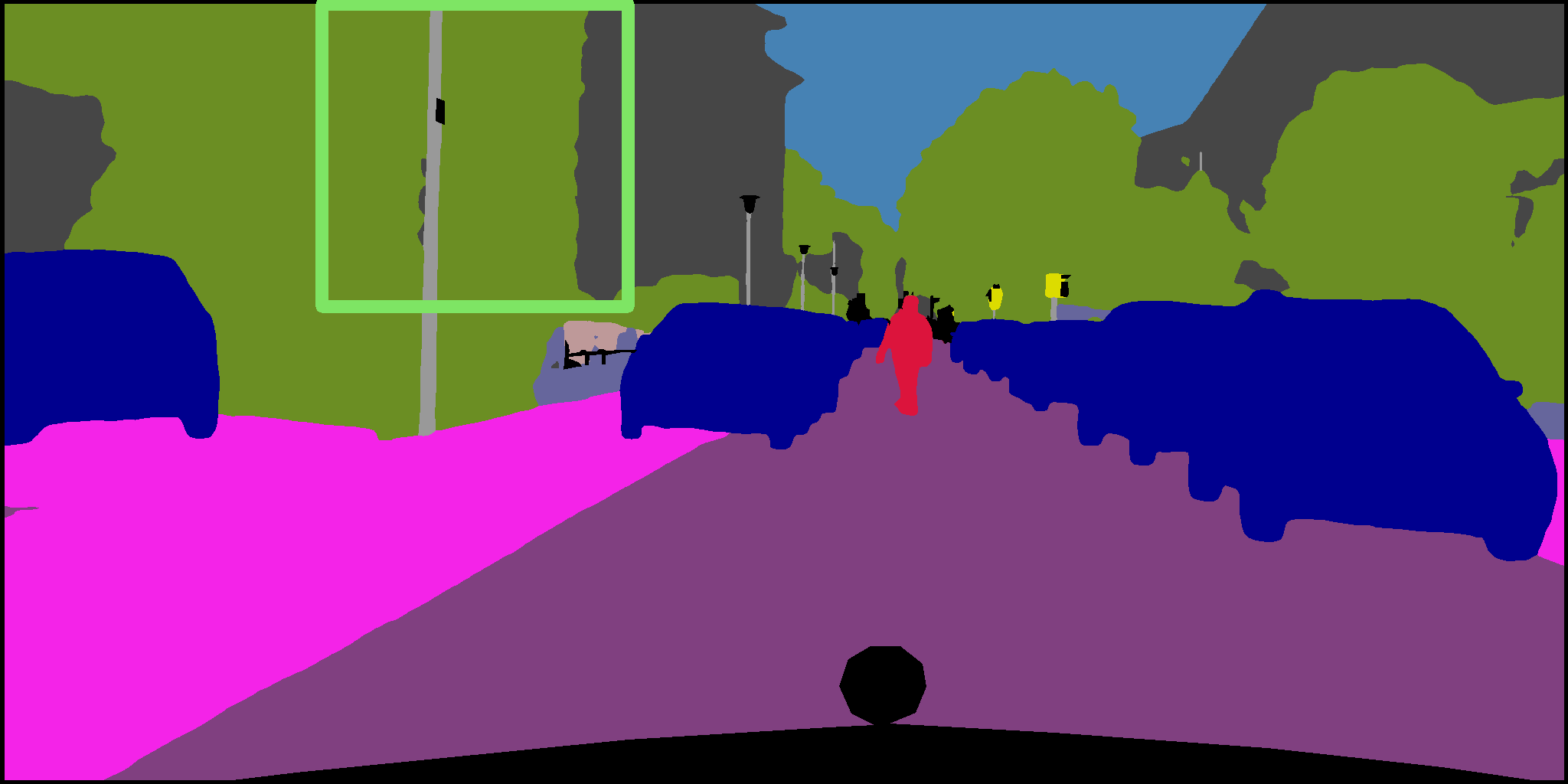} \\
        & \includegraphics[width=0.15\textwidth]{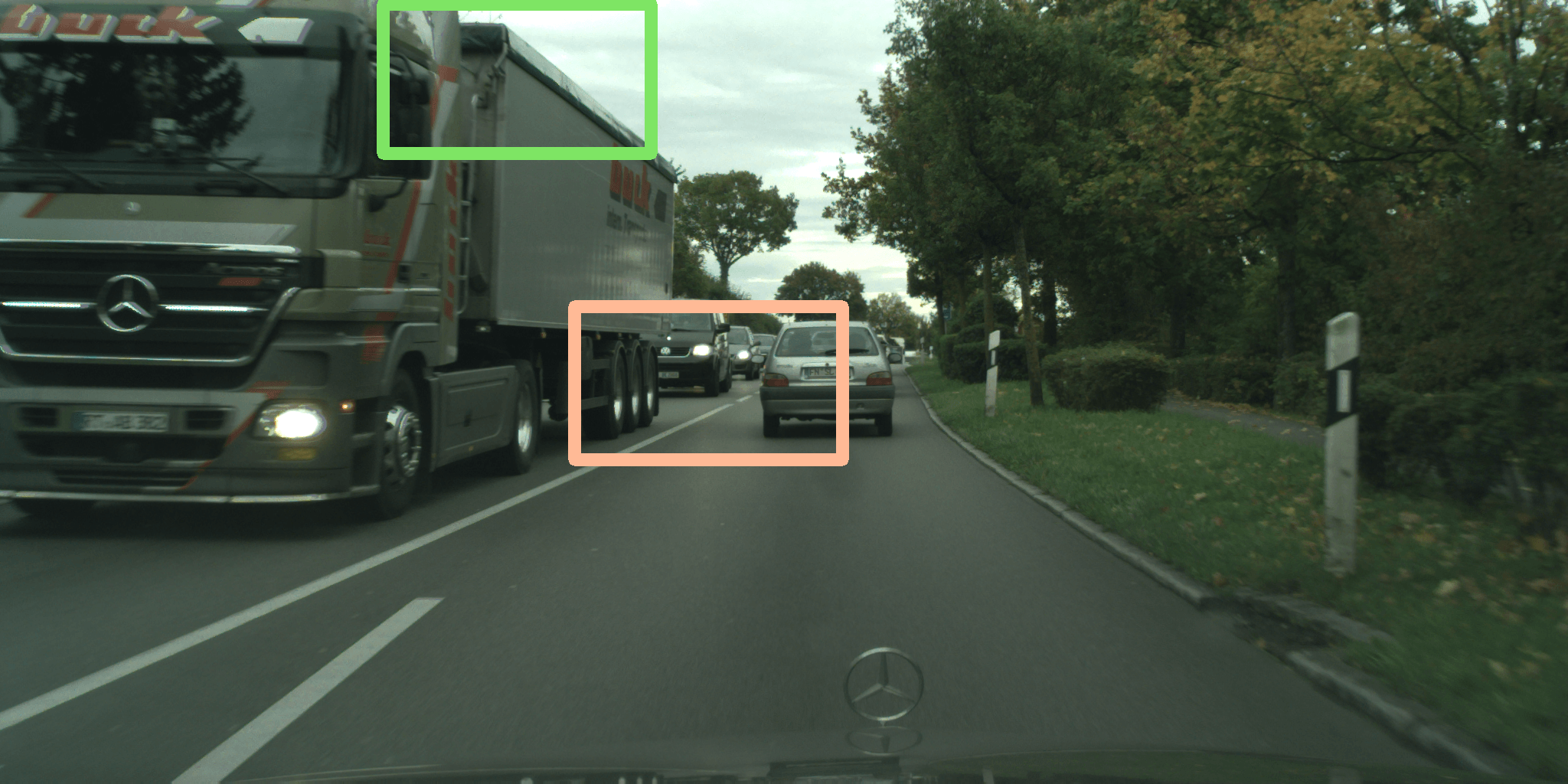} & 
        \includegraphics[width=0.15\textwidth]{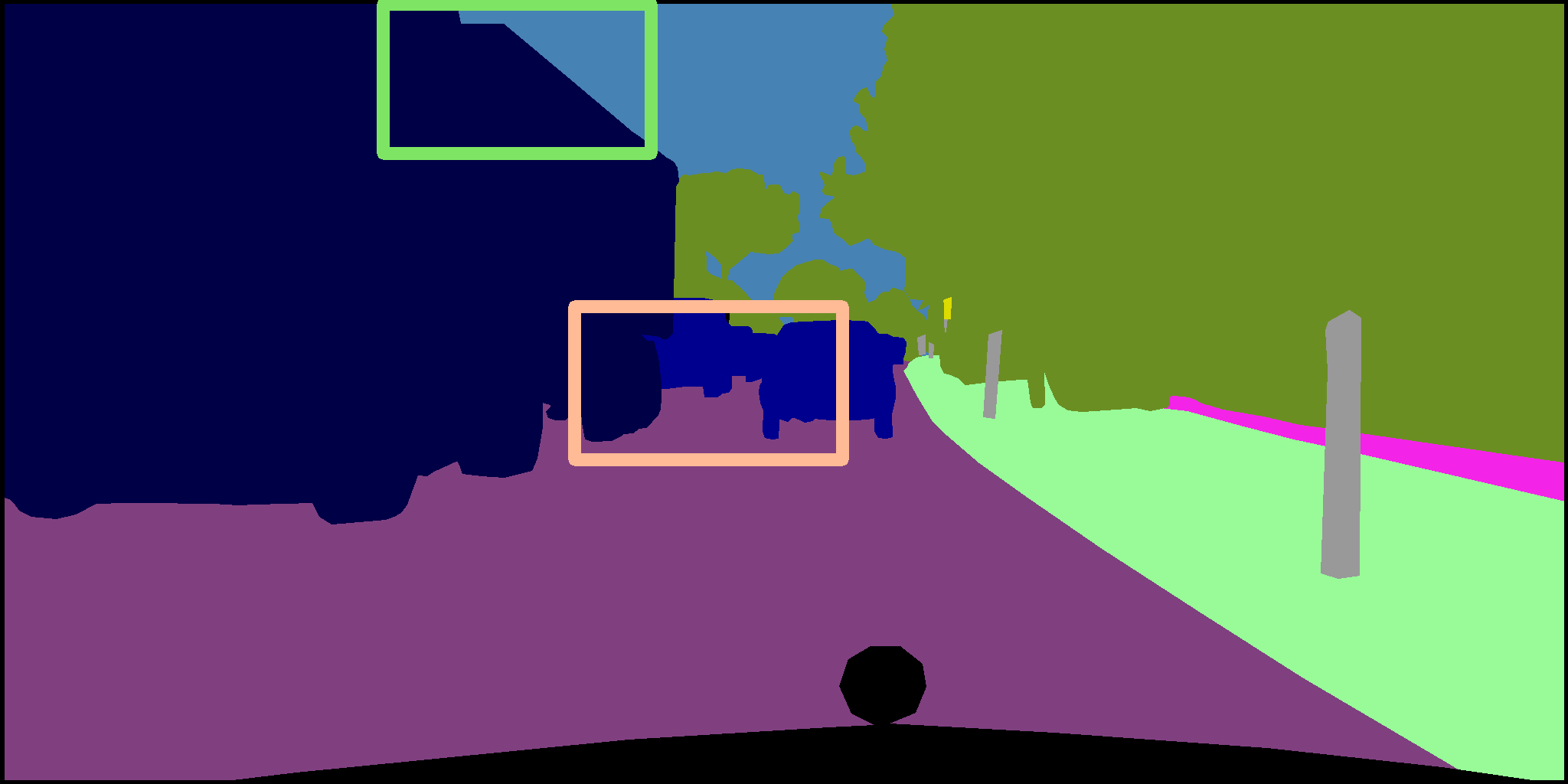} &
        \includegraphics[width=0.15\textwidth]{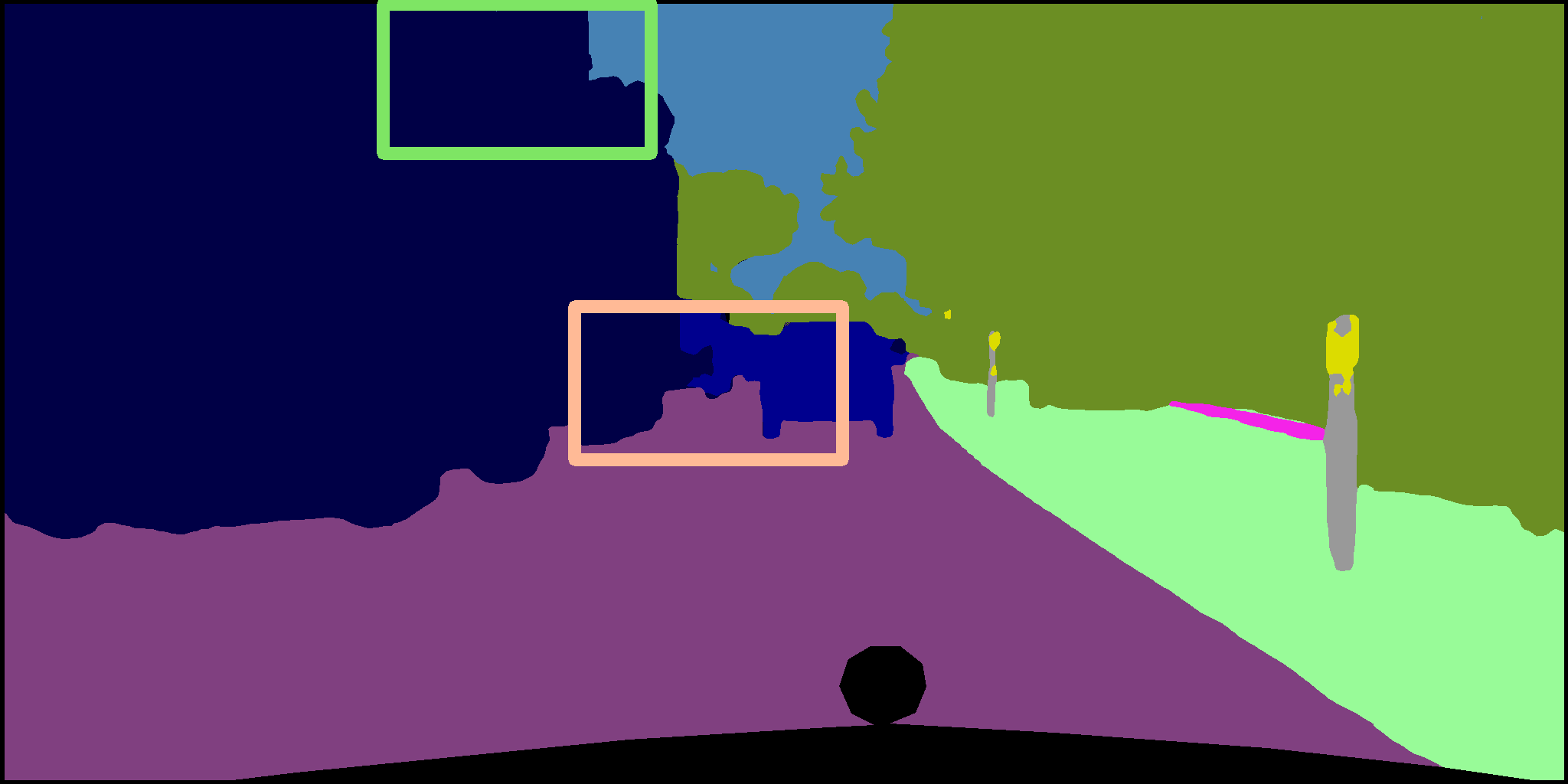} &
        \includegraphics[width=0.15\textwidth]{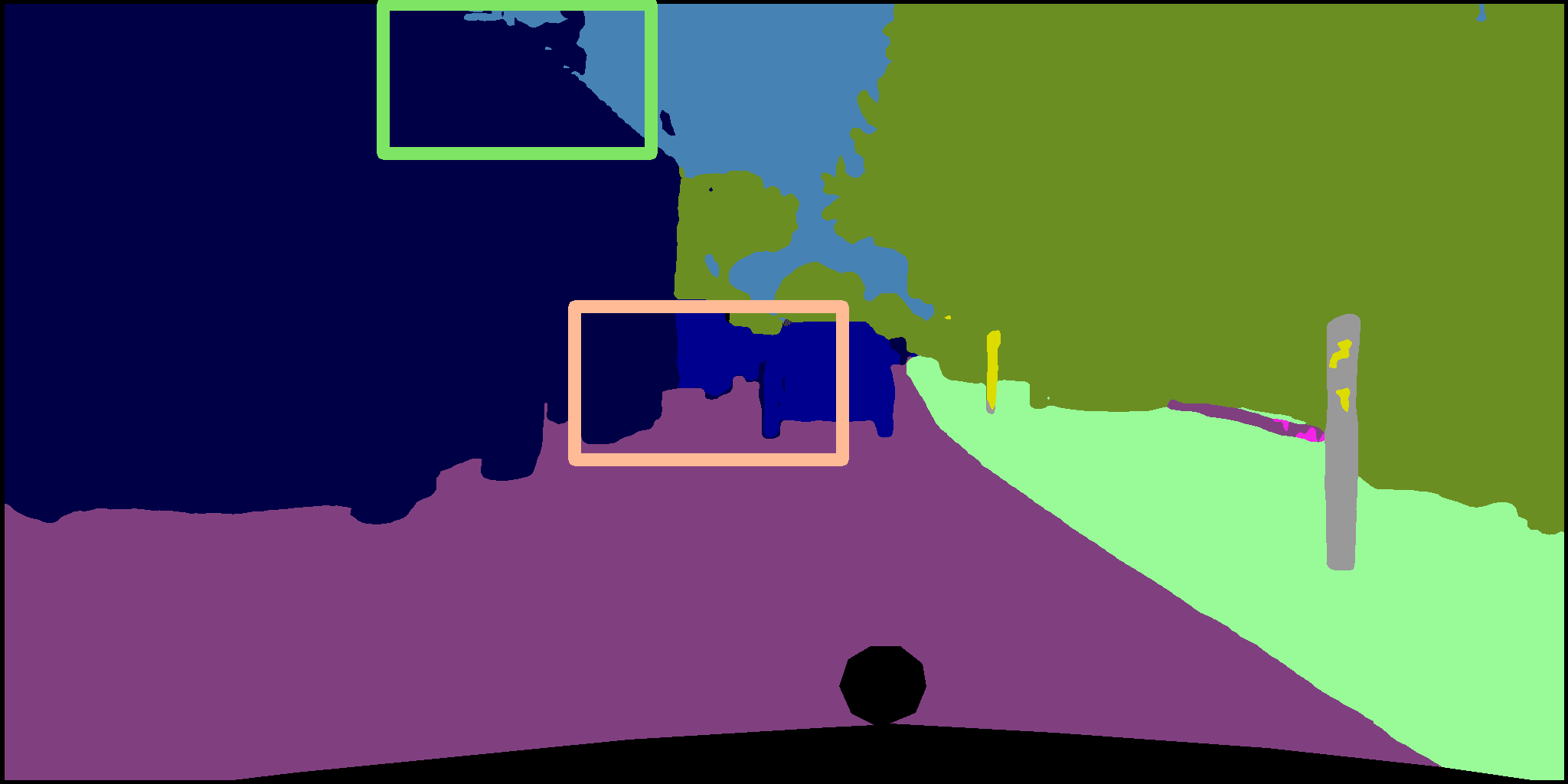} &
        \includegraphics[width=0.15\textwidth]{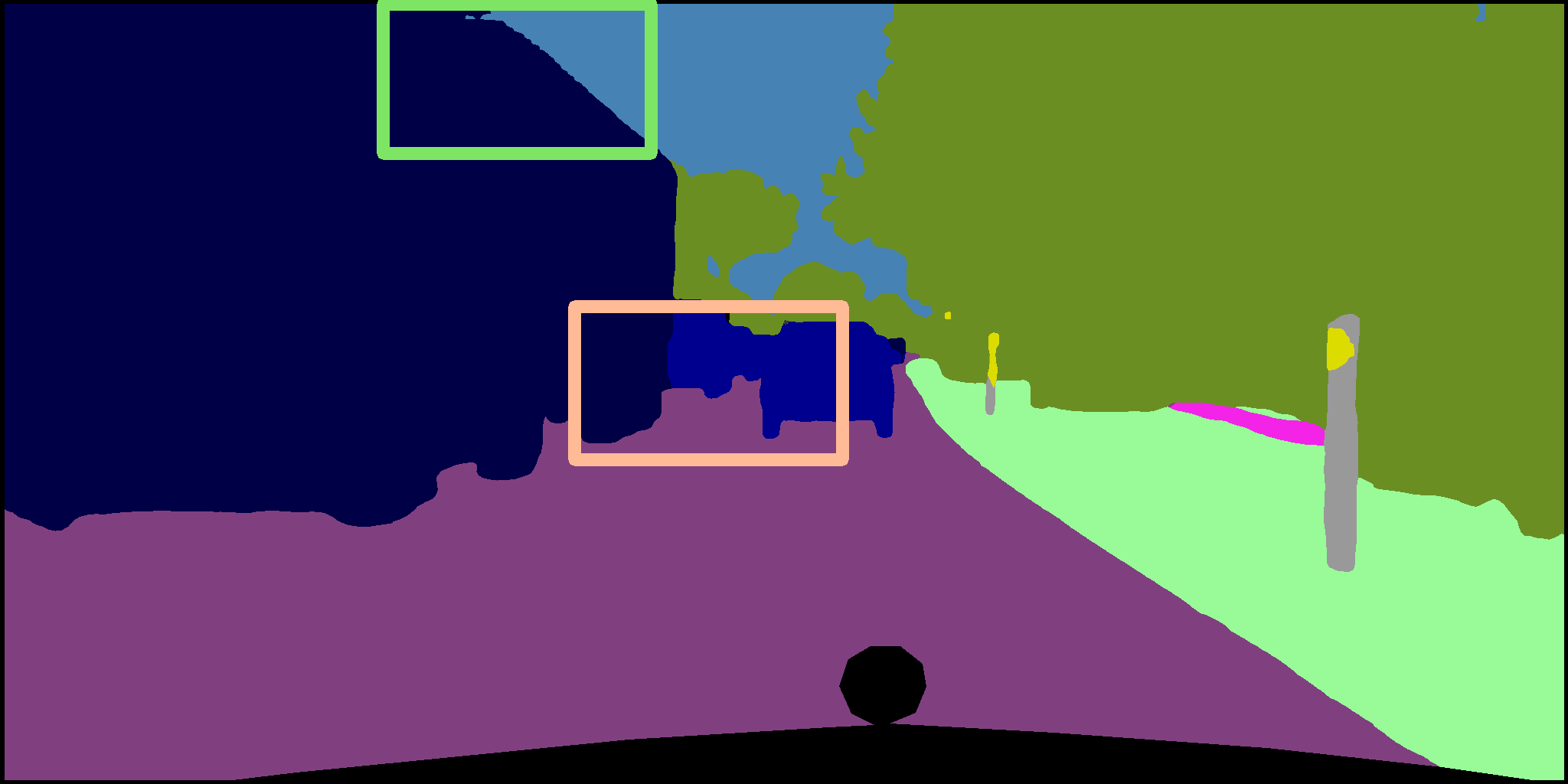} \\

        \multirow{2}{*}{\rotatebox{90}{ADE20K}} &
        \includegraphics[width=0.15\textwidth]{exp_imgs/a2w_val/rgb_hrocr_ADE_val_00000156_box.png} &
        \includegraphics[width=0.15\textwidth]{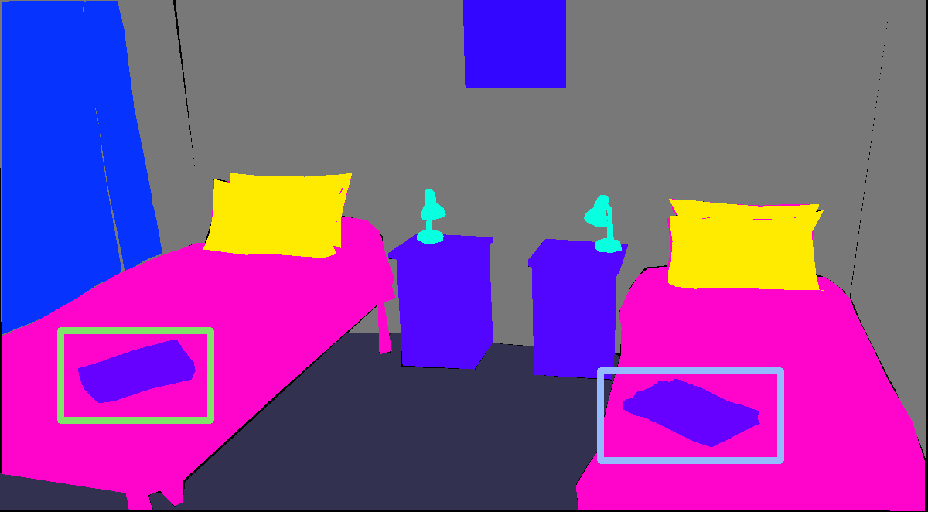}  &
        \includegraphics[width=0.15\textwidth]{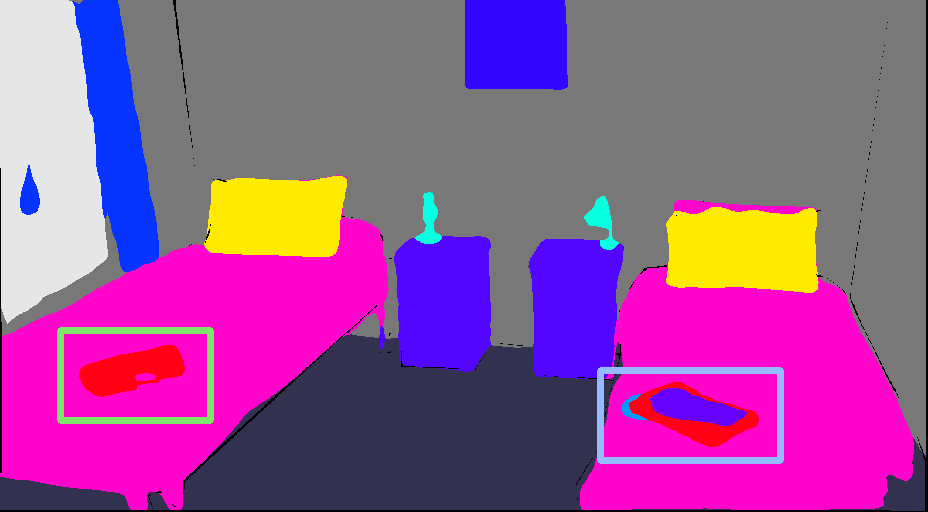}  &
        \includegraphics[width=0.15\textwidth]{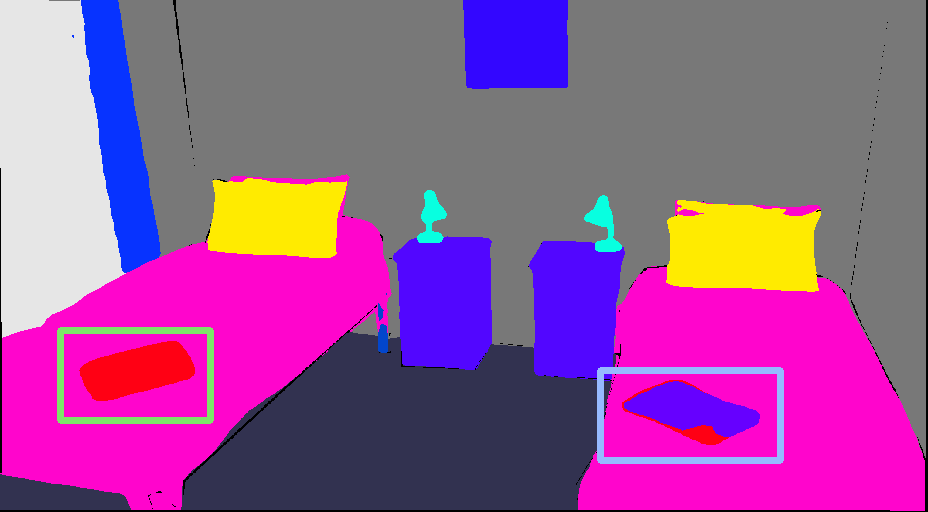}  &
        \includegraphics[width=0.15\textwidth]{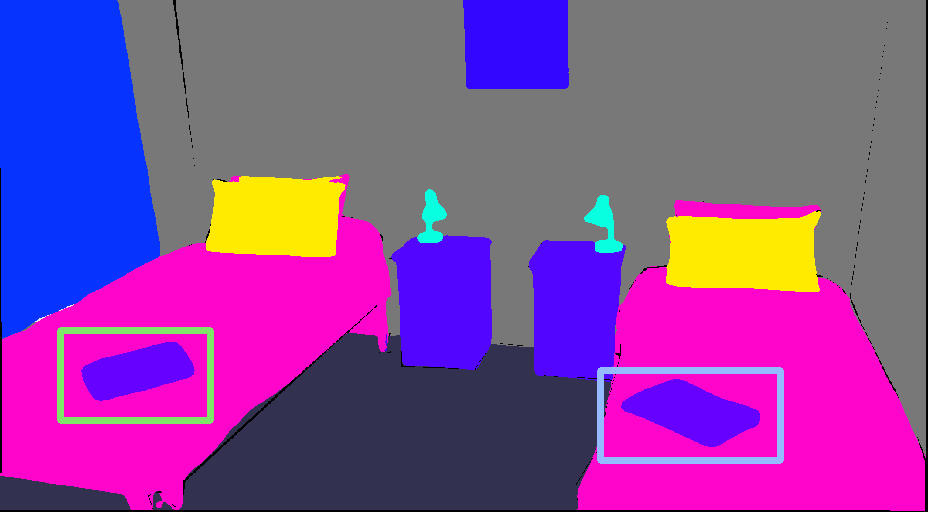} \\
        
        & \includegraphics[width=0.15\textwidth]{exp_imgs/a2w_val/rgb_hrocr_ADE_val_00000715_box.png}  &
        \includegraphics[width=0.15\textwidth]{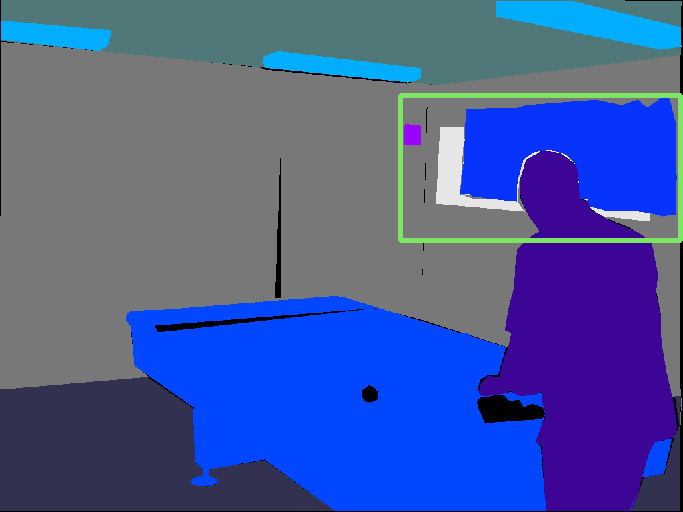}  &
        \includegraphics[width=0.15\textwidth]{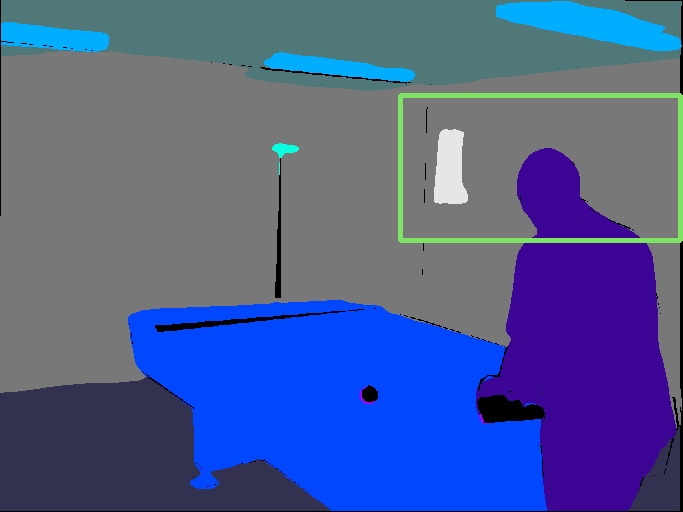}  &
        \includegraphics[width=0.15\textwidth]{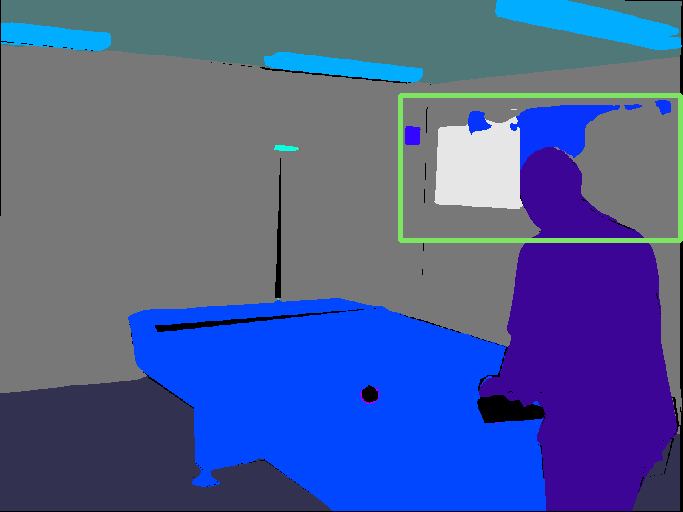}  &
        \includegraphics[width=0.15\textwidth]{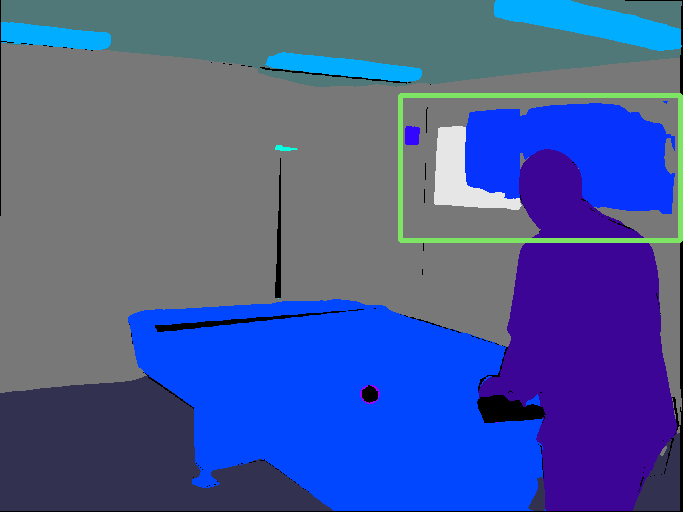} \\
        & RGB & GT & CE & CE+IoU & CE+IABL \\ 
        \end{tabular}

    }
\end{center}
\vspace{-10pt}
\caption{Qualitative results taken from Cityscapes \aaaiadd{and ADE20K} validation set.}
\label{fig:cs_detail}
\vspace{-12pt}
\end{figure}


\subsection{Ablation Studies}
\noindent \textbf{Loss terms.} \blue{We first test the influence of loss terms on the Cityscapes validation dataset by re-training the DeepLabV3 network and show the results in Tab.~\ref{tab:ablation_detach}. Since the gradient of ABL is not useful when PDBs are far from the GTBs, adding ABL at the beginning of the training does not improve network performance. Thus, we start to add ABL at the last $20$\% epochs to verify its effect, but only obtain a $0.1$\% improvement over mIoU. Then, we re-train the network with CE+IoU and CE+IABL. It shows that adding ABL to CE+IoU in training can increase the mIoU by $0.3$\%, and the combination of IoU loss and ABL, \ie CE+IABL contribute $1$\% improvement on mIoU in this study. Although the ABL does not contribute most to mIoU in this case, we do see the obvious improvement of boundary alignment in qualitative comparisons. In Tab.~\ref{tab:ablation_cie}, we test the contribution of each loss term on ADE20K dataset by re-training the OCR network. While the mIoU and pixel accuracy can both be improved after adding IoU loss and ABL, CE+IABL contributes most of the improvement to mIoU by around $0.65$\% over CE+IoU in the single-scale inference, which verifies the contribution of the proposed ABL in this experiment. We argue that the ABL can contribute more to a dataset with a large number of semantic classes and hence more GTBs. For instance, ADE20K has 150 classes, while Cityscapes only has 19 classes. More GTBs give the ABL more space to adjust the network's behavior.}


\noindent \textbf{Detaching operation.} We verify the effectiveness of the detaching operation in Tab.~\ref{tab:ablation_detach}. Significant drops of pixel accuracy and mIoU can be observed when training without the aforementioned detaching operation to suppress conflicts. Hence, \blue{it is important to control the gradient flow when there exist contradictory targets for KL divergence between two neighboring pixels.}

\noindent \textbf{FKL loss.} In Tabs.~\ref{tab:ablation_cie} and \ref{tab:compare_stm}, it can be seen that the combination of IoU loss and FKL, denoted by IFKL in the \nth{3} row, can also improve the pixel accuracy and mIoU quantitatively. However, CE+IFKL does \emph{not} perform as well as CE+IABL. We speculate that it is because FKL treats every pixel equally, while the ABL pays more attention to the pixels on the PDB. Such design allows the network to adjust its behavior in a progressive way, avoiding over-confident decisions when updating the network parameters. 

\noindent \tipadd{\textbf{Boundary pixels number threshold.}
We evaluate the influence of different thresholds on the Cityscapes validation set with the FCN~[backbone: HRNetV2-W18s]. The results are as follows: mIoU 75.59\% using 1\% threshold, 75.46\% using 2\%, and 75.41\% using 0.5\%. This empirically verifies that our choice of 1\% threshold is reasonable.}

\noindent \tipadd{\textbf{The degree of ABL's dependence on IoU loss.}
To evaluate ABL's contribution further, we design an \emph{IoU weight decay experiment}, which linearly decreases the weight of IoU loss from 1 to 0 during training but increase the weight of ABL from 0 to 1. It achieves mIoU 75.65\% on the Cityscapes validation set with the FCN~[backbone: HRNetV2-W18s], comparable to the mIoU 75.59\% trained with CE+IoU+ABL without weight decay. It can be seen that the decreased IoU weight does not lead to the downgrade of segmentation performance. Moreover, we do observe that ABL can refine the semantic boundary for thin structures and complex boundaries, as shown in Tab.~\ref{tab:BoundaryFscoreCategoryMS} and Figs.~\multiref{fig:compare_CE}{fig:vos_detail}.}

\begin{table*}[t]
    \small
    \centering
    \vspace{-1.5em}
    \newlength\savewidth
    \newcommand\shline{\noalign{\global\savewidth\arrayrulewidth
                               \global\arrayrulewidth 1.2pt}%
                      \hline
                      \noalign{\global\arrayrulewidth\savewidth}}
    \resizebox{0.85\linewidth}{!}{
    \begin{threeparttable}
        \begin{tabularx}{1.2\linewidth}{r|qqqqqqqqqqqqqqqqqqq@{\hspace{0.2in}}|Z}
        
            \shline
             method & \tabincell{l}{road} & \tabincell{l}{side-\\[-2pt]walk} & \tabincell{l}{buil-\\[-2pt]ding} & \tabincell{l}{wall} & \tabincell{l}{fence} & \tabincell{l}{pole} & \tabincell{l}{\textit{traffic}\\[-2pt]\textit{{light}}} & \tabincell{l}{traffic\\[-2pt]sign} & \tabincell{l}{vege-\\[-2pt]tation} & \tabincell{l}{terr-\\[-2pt]ian} & \tabincell{l}{sky} & \tabincell{l}{per-\\[-2pt]son} & \tabincell{l}{rider} & \tabincell{l}{car} & \tabincell{l}{truck} & \tabincell{l}{bus} & \tabincell{l}{train} & \tabincell{l}{moto-\\[-2pt]rcycle} & \tabincell{l}{bicy-\\[-2pt]cle} & mean\\
            \hline
             DeepLabV3 & 98.4 & 86.5 & 93.1 & 63.9 & 62.6 & 66.1 & \textit{72.2} & 80.0 & 92.8 & 66.3 & 95.0 & 83.3 & 65.5 & 95.3 & 74.5 & 89.0 & 80.0 & 67.4 & 78.4 & 79.5 \\
             +Segfix & \textbf{98.5} & \textbf{87.1} & \textbf{93.5} & \textbf{64.6} & \textbf{63.1} & \textbf{69.0} & \textit{74.9} & \textbf{82.4} & \textbf{93.2} & \textbf{66.7} & \textbf{95.3} & \textbf{84.9} & 66.9 & \textbf{95.8} & 75.0 & \textbf{89.6} & 80.7 & 68.4 & 79.7 & \textbf{80.5~($\uparrow$1.0)} \\
             {+IABL} & 98.1 & 84.9 & 93.1 & 59.9 & \textbf{63.1} & 68.3 & \textbf{\textit{75.4}} & 82.3 & 92.7 & 64.6 & 95.0 & 84.6 & \textbf{69.4} & 95.6 & \textbf{79.9} & 87.8 & \textbf{83.3} & \textbf{70.6} & \textbf{80.2} & \textbf{80.5~($\uparrow$1.0)} \\
            \hline
             OCR & 98.4 & 86.8 & 93.3 & 62.2 & 66.0 & 70.0 & \textit{73.9} & 82.0 & 93.0 & 67.2 & 95.0 & 84.3 & 66.3 & 95.6 & 82.9 & 91.7 & 85.0 & 67.7 & 79.2 & 81.1 \\
             +Segfix & \textbf{98.5} & \textbf{87.3} & 93.5 & 62.6 & \textbf{66.4} & \textbf{71.4} & \textit{75.7} & \textbf{83.3} & \textbf{93.3} & 67.6 & \textbf{95.3} & 85.2 & 67.2 & \textbf{96.0} & \textbf{83.3} & 92.2 & 85.5 & 68.6 & 80.0 & 81.7~($\uparrow$0.6) \\
             {+IABL} & 98.3 & 86.7 & \textbf{93.6} & \textbf{63.9} & 64.9 & 70.2 & \textbf{\textit{76.9}} & 83.2 & 93.2 & \textbf{69.0} & \textbf{95.3} & \textbf{85.6} & \textbf{70.7} & \textbf{96.0} & 80.0 & \textbf{93.3} & \textbf{86.6} & \textbf{69.3} & \textbf{80.8} & \textbf{82.0~($\uparrow$0.9)} \\
            \shline
        \end{tabularx}
    \end{threeparttable}
    }
    \vspace{-7pt}
    \caption{Class-wise mIoU results \aaaiadd{obtained using single-scale inference} on the Cityscapes validation set. \emph{+Segifx} indicates that Segfix is use to refine the baseline output. \emph{{+IABL}} indicates that the network is trained with additional loss IABL. \aaaidel{SS: single-scale inference. MS: multi-scale inference.}}
    \label{tab:mIoU_big}
    \vspace{-7pt}
\end{table*}

\begin{table*}[ht]
    \small
    \newlength\savewidthh
    \newcommand\shline{\noalign{\global\savewidthh\arrayrulewidth
                               \global\arrayrulewidth 1.2pt}%
                      \hline
                      \noalign{\global\arrayrulewidth\savewidthh}}
    \centering
    \resizebox{0.85\linewidth}{!}{
    \begin{threeparttable}
        \begin{tabularx}{1.2\linewidth}{S|D|AAAAAAAAAAAAAAAAAAA@{\hspace{0.2in}}|F}
            \shline 
              & scale & \tabincell{l}{road} & \tabincell{l}{side-\\[-2pt]walk} & \tabincell{l}{buil-\\[-2pt]ding} & \tabincell{l}{wall} & \tabincell{l}{fence} & \tabincell{l}{pole} & \tabincell{l}{\textit{traffic}\\[-2pt]\textit{{light}}} & \tabincell{l}{traffic\\[-2pt]sign} & \tabincell{l}{vege-\\[-2pt]tation} & \tabincell{l}{terr-\\[-2pt]ian} & \tabincell{l}{sky} & \tabincell{l}{per-\\[-2pt]son} & \tabincell{l}{rider} & \tabincell{l}{car} & \tabincell{l}{truck} & \tabincell{l}{bus} & \tabincell{l}{train} & \tabincell{l}{moto-\\[-2pt]rcycle} & \tabincell{l}{bicy-\\[-2pt]cle} & mean\\\hline
             \multirow{3}{*}{1px}  & OCR & 74.1 & 50.2 & 57.2 & 56.8 & 54.6 & 61.0 & \textit{64.8} & 60.6 & 55.6 & 51.3 & 65.7& 56.3 & 65.9 & 64.7 & 84.3 & 89.8 & 96.8 & 77.3 & 56.1 & 65.4 \\
             & +Segfix & \textbf{76.0} & \textbf{52.6} & \textbf{59.3} & \textbf{58.1} & \textbf{55.5} & \textbf{64.2} & \textit{67.8} & 64.1 & \textbf{57.7} & \textbf{52.9} & 67.1 & 59.3 & 67.6 & \textbf{68.7} & \textbf{84.7} & 89.8 & \textbf{97.0} & 78.4 & 58.6 & \textbf{67.3~($\uparrow$1.9)} \\
             & {+IABL} & 74.6 & 51.0 & 57.4 & 53.7 & 50.8 & 62.1 & \textbf{\textit{74.7}} & \textbf{65.1} & 55.7 & 52.3 & \textbf{66.8} & \textbf{60.1} & \textbf{68.9} & 67.0 & 84.6 & \textbf{90.3} & 96.4 & \textbf{79.5} & \textbf{61.5} & 67.0~($\uparrow$1.6) \\\hline
             \multirow{3}{*}{3px} & OCR & 86.5 & 70.1 & 75.7 & 62.5 & 60.1 & 79.6 & \textit{77.8} & 78.9 & 76.4 & 58.6 & 82.9 & 73.9 & 76.7 & 84.6 & 86.5 & 92.8 & 97.4 & 80.4 & 71.3 & 77.5 \\
              & +Segfix & \textbf{87.2} & \textbf{71.0} & \textbf{76.4} & \textbf{63.0} & \textbf{60.7} & 79.7 & \textit{78.5} & 79.3 & \textbf{77.3} & 60.0 & 83.5 & 74.5 & 77.4 & 85.9 & 86.6 & 92.3 & \textbf{97.5} & 81.1 & 72.3 & 78.1~($\uparrow$0.6) \\
              & {+IABL} & 86.2 & 70.1 & 75.4 & 59.4 & 56.3 & \textbf{80.7} & \textbf{\textit{86.9}} & \textbf{81.9} & 76.5 & \textbf{60.1} & \textbf{84.1} & \textbf{77.6} & \textbf{80.4} & \textbf{86.1} & \textbf{87.2} & \textbf{93.2} & 97.0 & \textbf{83.1} & \textbf{76.9} & \textbf{78.9~($\uparrow$1.4)} \\\hline
             \multirow{3}{*}{5px} & OCR & 90.3 & 76.4 & 82.3 & 65.0 & 62.9 & \textbf{82.8} & \textit{81.7} & 82.6 & 84.4 & 62.1 & 88.1 & 78.9 & 80.8 & 89.5 & 87.5 & 93.7 & \textbf{97.7} & 81.9 & 77.7 & 81.4 \\
              & +Segfix & \textbf{90.6} & \textbf{76.9} & \textbf{82.5} & \textbf{65.1} & \textbf{63.0} & 82.6 & \textit{81.8} & 82.3 & \textbf{84.6} & 63.1 & 88.2 & 78.7 & 81.1 & 90.1 & 87.3 & 93.1 & \textbf{97.7} & 82.3 & 77.8 & 81.5~($\uparrow$0.1) \\
              & {+IABL} & 89.8 & 76.4 & 81.8 & 61.7 & 58.6 & 83.8 & \textbf{\textit{89.9}} & \textbf{84.9} & 84.3 & \textbf{63.3} & \textbf{89.1} & \textbf{82.2} & \textbf{84.1} & \textbf{90.6} & \textbf{88.1} & \textbf{94.0} & 97.2 & \textbf{84.6} & \textbf{82.5} & \textbf{82.5~($\uparrow$1.1)}\\ 
            \shline 
        \end{tabularx}
    \end{threeparttable}
    }
    \vspace{-7pt}
    \caption{Class-wise Boundary F-score results \aaaiadd{obtained using multi-scale inference} on the Cityscapes validation set. \aaaidel{We utilize the OCR network \aaaiadd{and multi-scale inference} in these experiments. \emph{+Segifx} indicates that Segfix is use to refine the baseline output. \emph{{+IABL}} indicates that the network is trained with additional loss IABL. \aaaidel{SS: single-scale inference. MS: multi-scale inference.}}}
    \label{tab:BoundaryFscoreCategoryMS}
    \vspace{-10pt}
\end{table*}


\subsection{Quantitative Evaluation}

\noindent \textbf{Results on ADE20K and Cityscapes validation sets.} In Tabs.~\ref{tab:ade20k_big} and \ref{tab:cityscapes_big}, we show that training with IABL along with cross-entropy loss can improve the pixel accuracy and mIoU over state-of-the-art image segmentation networks. As for the ADE20K dataset, training the OCR network \cite{cordts2016cityscapes} with additional IABL improves the mIoU and pixel accuracy by $1.22$\% and $0.23$\% over that trained with cross-entropy only on the validation set\aaaiadd{. Not only effective on CNN-based network, additional IABL supervision on Transformer-based network UperNet \cite{xiao2018unified} [backbone: SwinTransformer-B \cite{DBLP:journals/corr/abs-2103-14030}] also brings an improvement in the mIoU by $0.74$\% on the validation set, which ranks the first place in this table (Tab.~\ref{tab:ade20k_big}, last row). }Similarly, \aaaiadd{for the Cityscape dataset,} training the OCR network with IABL\aaaidel{ on the Cityscape dataset} improves the mIoU by $0.7$\% on the validation set. 

\begin{figure}[t]
\begin{center}
    \setlength{\tabcolsep}{1pt}
    \resizebox{0.9\linewidth}{!}{
        \begin{tabular}{ccccc}
        \vspace{-3pt}
        \rotatebox{90}{\ \ \ \ \  \small{\#25}} &      
        \includegraphics[width=0.3\linewidth]{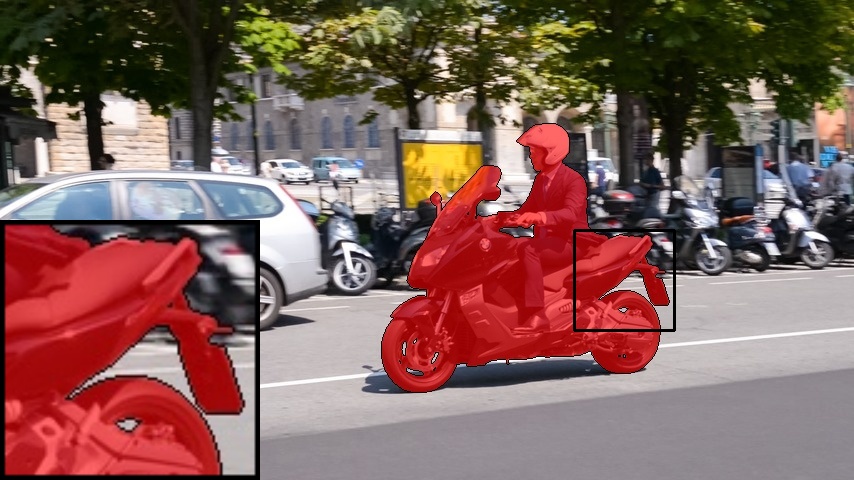} \hspace{-1mm} & 
        \includegraphics[width=0.3\linewidth]{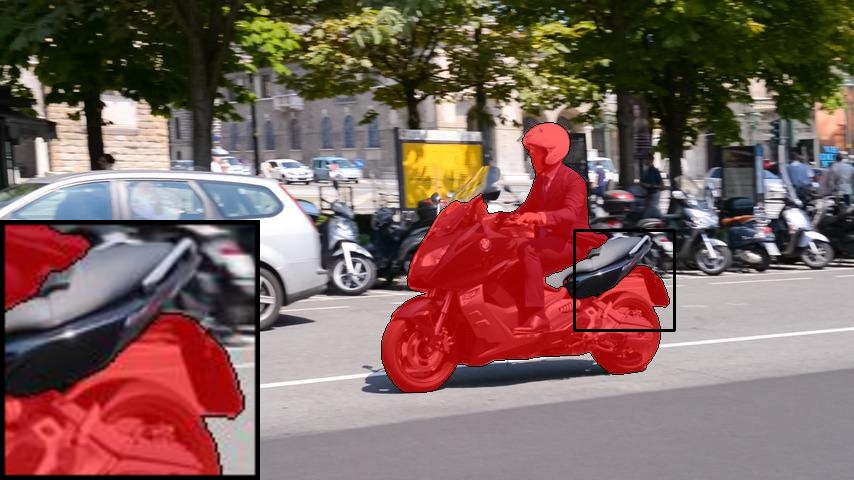} \hspace{-1mm} & 
        \includegraphics[width=0.3\linewidth]{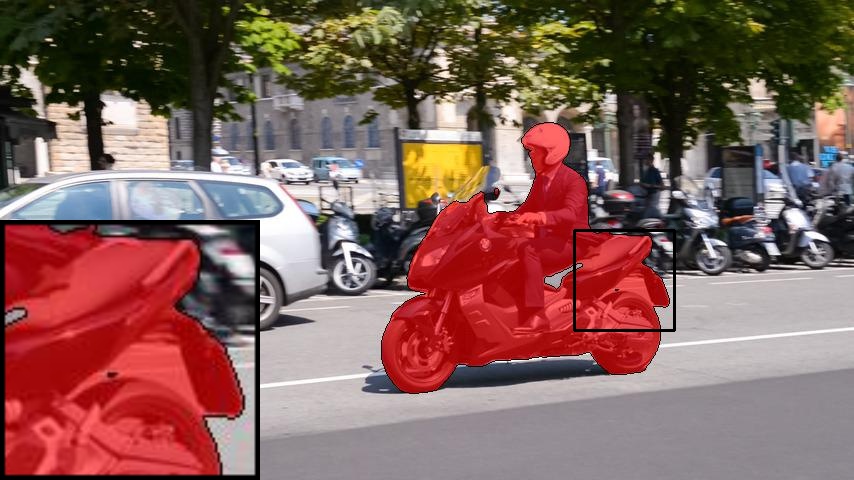} \hspace{-1mm} & 
        \includegraphics[width=0.3\linewidth]{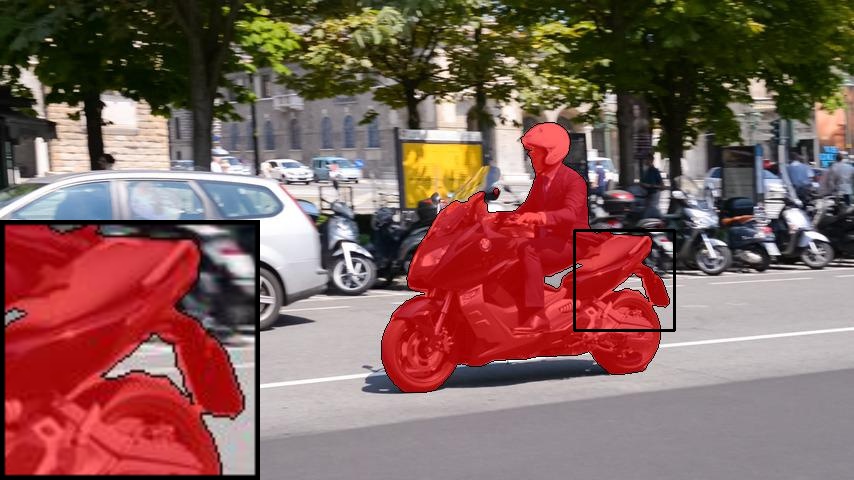} \hspace{-1mm} \\
        
        \vspace{-2pt}
        \rotatebox{90}{\ \ \ \ \  \small{\#26}} & 
        \includegraphics[width=0.3\linewidth]{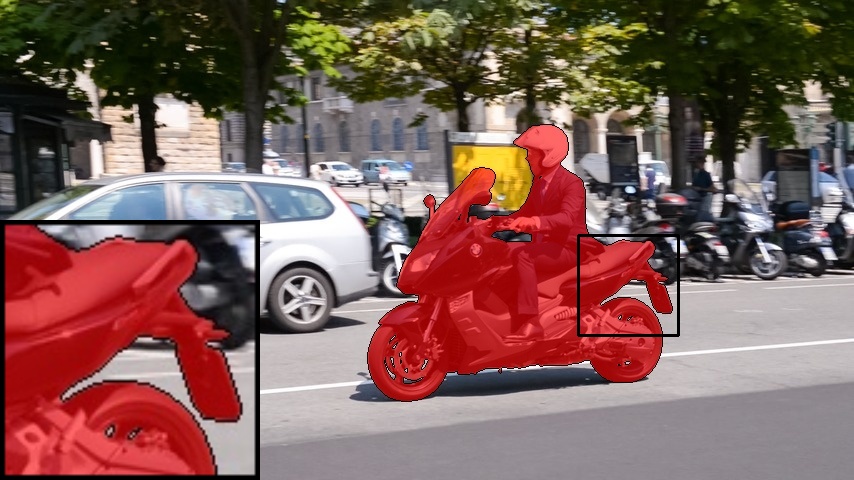} \hspace{-1mm} & 
        \includegraphics[width=0.3\linewidth]{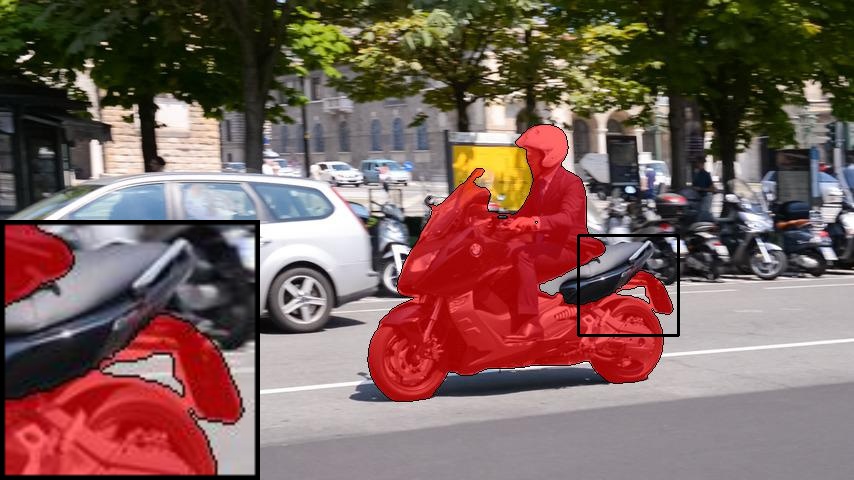} \hspace{-1mm} & 
        \includegraphics[width=0.3\linewidth]{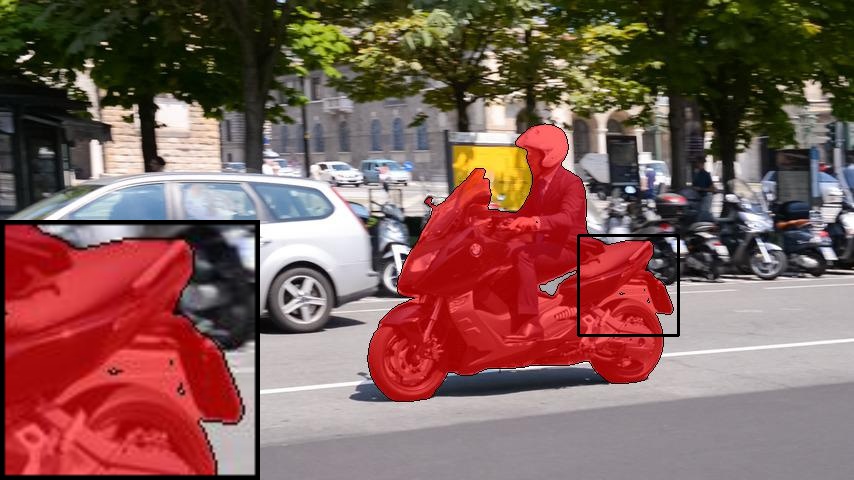} \hspace{-1mm} & 
        \includegraphics[width=0.3\linewidth]{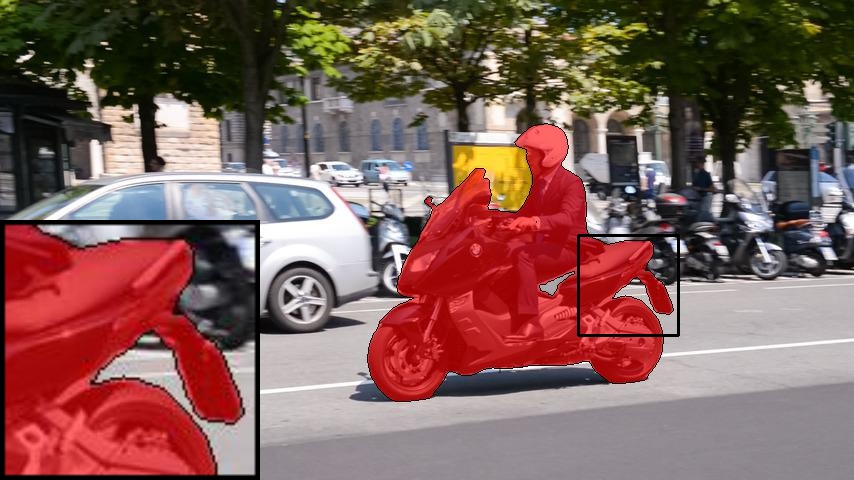} \hspace{-1mm} \\
        \vspace{-2pt}
        \rotatebox{90}{\ \ \ \ \  \small{\#27}} &        
        \includegraphics[width=0.3\linewidth]{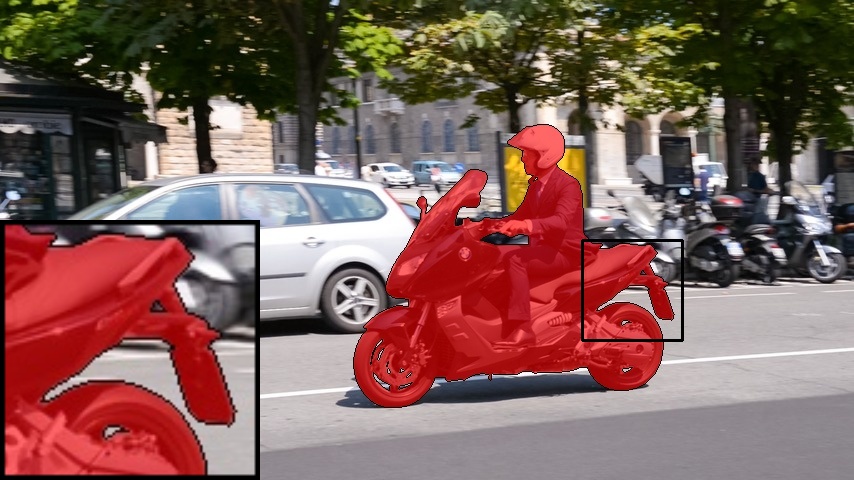} & 
        \includegraphics[width=0.3\linewidth]{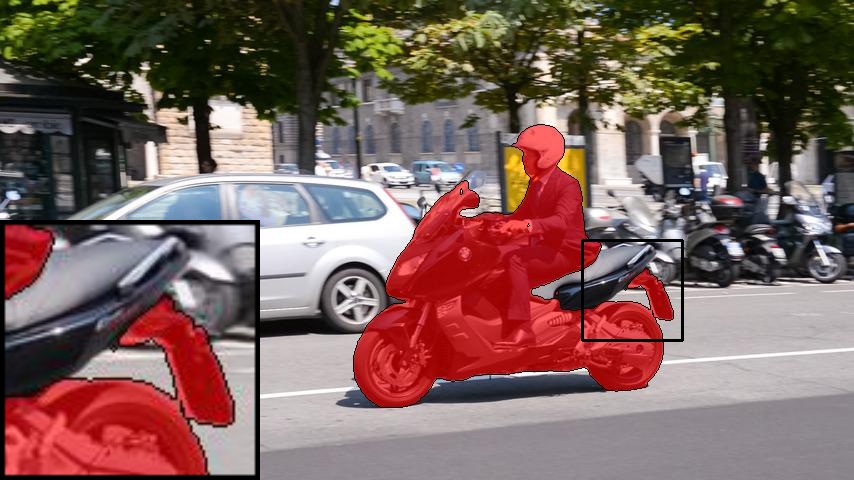} & 
        \includegraphics[width=0.3\linewidth]{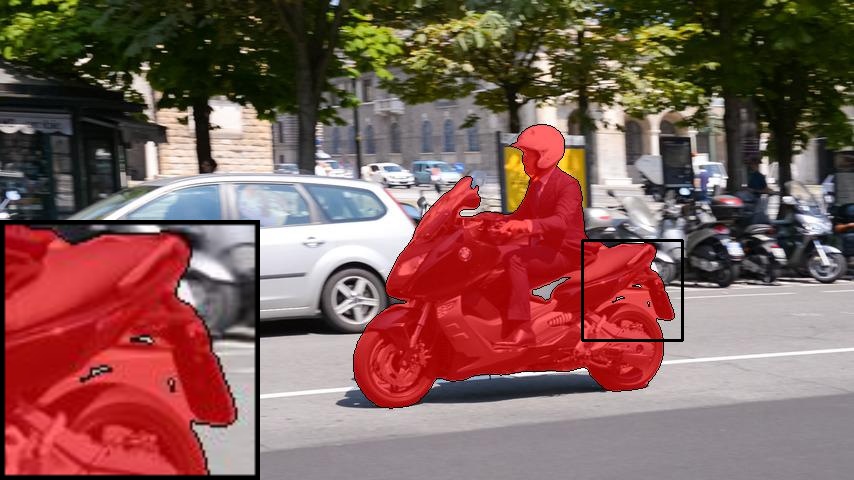} & 
        \includegraphics[width=0.3\linewidth]{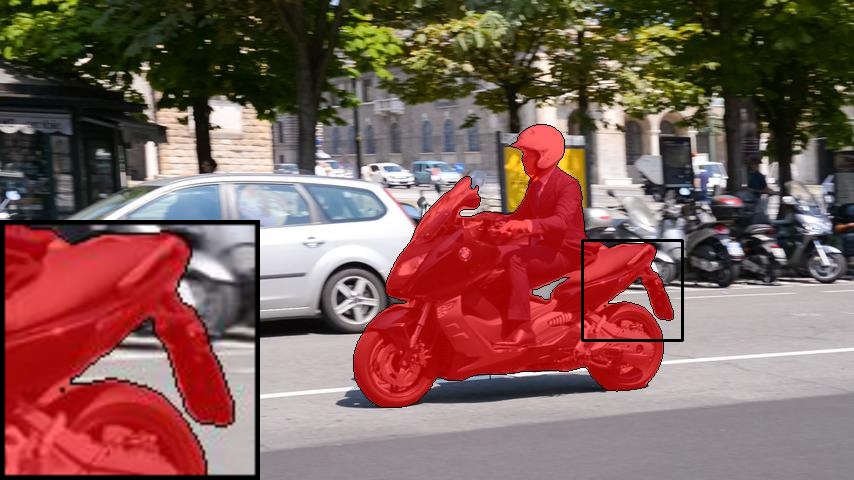} \\
        \vspace{-2pt}
        & \small{GT} \hspace{-1mm} & \small{STM} & \small{FT CE+IoU} \hspace{-1mm} & \small{FT CE+IABL} \hspace{-1mm} \\ 
        &  \multicolumn{4}{c}{scooter-black \aaaiadd{(43 frames in total)}}
        \end{tabular}
    }
\end{center}
\vspace{-12pt}
\caption{Qualitative results taken from DAVIS-2016 validation set. VOS Network: STM \cite{oh2019video}. FT: STM fine-tuning. \aaaidel{f: Total number of video frames. }\#N: frame number. The boundary details are more accurate after fine-tuning with additional loss IABL. \aaaidel{Best view on screen and zoom in.}}
\label{fig:vos_detail}
\vspace{-10pt}
\end{figure}

\noindent \textbf{Comparison with Segfix.} We compare our method with Segfix \cite{yuan2020segfix} on the Cityscapes validation set by using mIoU and boundary F-score metrics, since both methods focus on improving boundary details in semantic segmentation. In Tab.~\ref{tab:mIoU_big}, our method achieves comparable performance when using DeepLabV3 as the segmentation network, but improves the mIoU over Segfix by $0.3$\% when using the OCR network. In Tab.~\ref{tab:BoundaryFscoreCategoryMS}, we show the class-wise boundary F-scores of Segfix and our method. The scores are computed using the GTB dilation parameters $1$, $3$, and $5$ pixels. While Segfix outperforms in the cases of $1$ pixel, our method achieves a higher score in the cases of $3, 5$ pixels. 

\tipadd{Segfix is an elegant boundary refinement solution that propagates the interior labels to class boundaries. However, the propagation operation might downgrade the segmentation performance for thin objects that contain a small number of interior pixels. In contrast, the ABL is an end-to-end training loss that encourages the alignment of PDBs and GTBs, which achieves better mIoU and boundary F-scores, even for thin structures. Taking the class of traffic light as an example~(Tab~\ref{tab:BoundaryFscoreCategoryMS}, \nth{9} column)}, our method achieves a consistent improvement of the boundary F-score over Segfix in all parameter settings, which shows that our method can handle boundaries of thin objects well. Since Segfix is a post-processing method, it can also be used to improve the segmentation results of our method. 

\noindent\tipadd{\textbf{Comparison with Boundary Loss.} We train ABL + generalized Dice loss~(GDL) \cite{DBLP:conf/miccai/SudreLVOC17} on the white matter hyperintensities~(WMH) dataset with the same network architecture and training parameters as \cite{kervadec2019boundary} for a fair comparison. We also use the same learning rate 0.001, batch size 8, training epochs 200, and loss conjunction method: $\mathbf{Loss} = \alpha * \mathbf{GDL} + (1-\alpha) * \mathbf{ABL}$, where $\alpha$ linearly decreases from 1 to 0.01. In Tab.~\ref{tbl:vsbl}, we show that training with ABL + GDL achieves higher dice similarity coefﬁcient (DSC) and smaller Hausdorff distance (HD) than Boundary Loss~(BL) + GDL. Moreover, we extend BL to a multiple-class loss and make a further comparison on Cityscapes validation set. In Tabs.~\ref{tab:ablation_detach} and \ref{tab:ablation_cie}, IoU+ABL archieves higher mIoU than IoU+BL.}

\tipadd{The motivation of BL is to minimize the distance between GTBs and PDBs. With the geo-cuts optimization techniques \cite{boykov2006integral}, this problem is converted to minimize the regional integral. This behavior will weaken the influence of pixels near GTBs since the distance weights there are much smaller, and the ratio of these pixels is small compared to the image size. In contrast, ABL focuses on PDB pixels, which can achieve better alignment. BL needs to work with a region-based IoU loss \aaaiadd{GDL}\aaaidel{, generalized Dice loss}, to avoid making the network collapse quickly into empty foreground classification results. Similarly, we use ABL and IoU loss together.}


\begin{table}[t]
	\renewcommand\arraystretch{0.9}
    \small
    \centering
    \resizebox{0.9\linewidth}{!}{
    \begin{tabularx}{\linewidth}{l@{\hspace{0.08in}}c@{\hspace{0.12in}}c@{\hspace{0.12in}}ccc}
         \toprule[1pt]
        \aaaiadd{Fine-tuning} & No & YES & YES & YES & YES \\
        Loss & CE & CE & CE+IoU & CE+IFKL & CE+IABL \\
         \midrule
        $\mathcal{J}$-mean & 88.67 & 88.81 & 89.08 & 89.08 & \textbf{89.29} \\
        $\mathcal{F}$-mean & 89.86 & 90.25 & 90.66 & 90.63 & \textbf{90.82}\\
        \bottomrule[1pt]
        
    \end{tabularx}
    }
    \vspace{-7pt}
    \caption{VOS results on DAVIS-2016 \cite{Perazzi2016} validation set. \aaaiadd{VOS network: STM}. The definition of region similarity metric $\mathcal{J}$-mean and contour accuracy $\mathcal{F}$-mean can be found in the DAVIS-2016 dataset \cite{Perazzi2016}.}
    \label{tab:compare_stm}
    \vspace{-15pt}
\end{table}


\noindent \textbf{VOS results.} We fine-tune the state-of-the-art VOS network STM \cite{oh2019video} with our loss to verify that our method can also be applied to VOS. Specifically, the STM is fine-tuned for $1$k iterations with batch size $4$ on both DAVIS-2016 \cite{Perazzi2016} and YouTube-VOS \cite{xu2018youtube} training data. The learning rate is set to $5\mathrm{e}{-8}$, and the weight $w_{a}$ is set to $5.0$. In Tab.~\ref{tab:compare_stm}, it can be seen that fine-tuning with addition IABL can improve region similarity metric $\mathcal{J}$-mean and contour accuracy $\mathcal{F}$-mean by around $0.7$\% and $1$\%, respectively, when testing on Davis-2016 validation set. 
Similar to image segmentation, training with CE+IABL can improve over CE+IoU loss, which also verifies ABL's contribution in VOS. However, adding FKL loss does not show superior performance, as shown in the \nth{5} colume of Tab.~\ref{tab:compare_stm}.

 \begin{table}[t]
	 \renewcommand\arraystretch{0.85}
     \small
     \centering
     \begin{tabular}{cccc}
         \toprule[1pt]
         Loss & GDL & GDL+BL & GDL+ABL \\
         \midrule
         DSC & 0.727 & 0.748 & \textbf{0.768} \\
         HD(mm) & 1.045 & 0.987 & \textbf{0.980} \\
        \bottomrule[1pt]     
     \end{tabular}
     \vspace{-7pt}
     \caption{BL vs. ABL on WMH validation set. \aaaidel{GDL: generalized Dice loss.}} 
     \label{tbl:vsbl}
     \vspace{-15pt}
 \end{table}

\subsection{Qualitative Results}
Fig.~\ref{fig:compare_CE} illustrates the progressive refinement of boundary details when using IABL as the additional training loss. This result is obtained when training DeepLabV3 on the Cityscapes dataset. It can be seen that the PDBs (red lines) of the traffic light and other objects are pushed toward the GTBs (blue lines). In Figs.~\ref{fig:teaser} and \ref{fig:cs_detail}, we show how adding loss terms influences the quality of semantic boundaries. The results show that the proposed ABL can greatly improve the semantic boundary details. Fig.~\ref{fig:vos_detail} illustrates the improved boundary details when fine-tuning STM with additional IABL. It also shows that fine-tuning with CE+IABL can further improve the boundary details over CE+IoU, such as the tail of motorcycle. 


\section{Conclusion}

In this work, we proposed an active boundary loss to be used in the end-to-end training of segmentation networks. Its advantage is that it allows the propagation of the ground-truth boundary information using a distance transform so as to regulate the network behavior at predicted boundaries. We have demonstrated that integrating the ABL into the network training can substantially improve the boundary details in semantic segmentation. 

In the future, it would be interesting to investigate how to reduce conflicts in our loss to further control the network behavior around boundaries efficiently. In addition, we plan to explore how to design boundary-aware loss to improve the boundary details in the task of depth prediction. 

\section{Acknowledgments}
\crc{We thank the reviewers for their constructive comments. Weiwei Xu is partially supported by NSFC~(No.~61732016).}

\fontsize{9.8pt}{10.8pt} \selectfont \bibliography{zbib}

\end{document}